\documentclass[11pt]{article}
\usepackage{booktabs}
\usepackage{graphicx}
\usepackage{placeins}
\usepackage[preprint]{acl}
\usepackage[utf8]{inputenc}
\usepackage{times}
\usepackage{titlesec}
\usepackage{xcolor}
\usepackage{latexsym}
\usepackage{float}
\usepackage{booktabs}
\usepackage{adjustbox}
\usepackage{multirow}
\usepackage{enumitem}
\usepackage{caption}
\usepackage{amsmath,bm}
\usepackage{amsfonts}
\usepackage{amssymb}
\usepackage{multirow}
\usepackage{lscape}   
\usepackage{CJKutf8}
\usepackage[T1]{fontenc}
\usepackage{microtype}
\usepackage{inconsolata}

\title{When Does Hate Transfer?
\\A Large-Scale Study of Cross-Lingual Hate Speech Detection }

\newcommand{\change}[2]{#1}

\author{
  Zahra Bokaei \\
  School of Informatics \\
  University of Edinburgh \\
  \texttt{zahra.bokaei@ed.ac.uk}
  \And
  Walid Magdy \\
  School of Informatics \\
  University of Edinburgh \\
  \texttt{wmagdy@ed.ac.uk}
  \And
  Bonnie Webber \\
  School of Informatics \\
  University of Edinburgh \\
  \texttt{bonnie.webber@ed.ac.uk}
}

\titlespacing\section{0pt}{8pt plus 2pt minus 2pt}{4pt plus 2pt minus 2pt}

\usepackage{graphicx}
\begin{document}
\maketitle
\begin{abstract}

Effective detection of hate content is essential for maintaining healthy online communities, yet performance can vary substantially across languages. This work investigates a core practical question: Can training data from one language support hate-speech detection in another language with comparable performance? To answer this question, we conduct a large-scale study across 14 languages. \change{We observe consistent patterns in which certain language pairs transfer more effectively than others. Among the factors we measure, linguistic similarity is the strongest predictor of these patterns, with temporal overlap and dataset-size difference playing smaller roles. These measurable factors do not fully explain the observed patterns; a complementary qualitative analysis suggests that shared cultural cues and communicative norms may be associated with some of the remaining cases, particularly within several clusters. We present this cultural interpretation as a hypothesis from qualitative inspection rather than a measured cause.}{} Our analysis includes low-resource languages, offering practical insights into when cross-lingual transfer can support them and where its limitations emerge. These findings inform the design of more reliable multilingual hate-detection systems.

\textbf{Warning:} This paper includes examples of hate-speech for research purposes.
\end{abstract}

\section{Introduction}

Hate speech detection aims to identify harmful expressions in online text, such as hate-speech, harassment, and threats \cite{hoang-etal-2024-toxcl}. As digital communication expands globally, exposure to such content has increased, motivating efforts to build automated systems capable of supporting safer online communities. However, what counts as hate-speech is context-dependent, and even human annotators vary in how they perceive it across settings \cite{sap-etal-2022-annotators}. These differences pose a challenge for developing models that generalize reliably across languages. 

Large Language Models (LLMs) provide strong multilingual representations and are increasingly used for hate detection \cite{khondaker-etal-2023-gptaraeval, kumar2024towards, abaskohi-etal-2024-benchmarking}. However, cross-lingual performance remains uneven, and the factors shaping transfer effectiveness are underexplored \cite{meyer-buys-2024-systematic, BOKAEI2026104912}. This motivates a practical question: under what conditions can training data from one language support hate-speech detection in another? Potential factors include dataset size, label definitions, temporal alignment, and linguistic similarity, while prior work suggests that social or regional context may also shape how hate expressions are formed and recognized \cite{bokaei-etal-2025-culture, pawar2024cultural}. We treat this last factor as a hypothesis to be explored qualitatively. In this study, we analyse hate-speech transfer across 14 languages from diverse regions and domains. We construct a cross-lingual transfer matrix by training classifiers on each language and evaluating them on all others. We then run two multilingual experiments: (1) integrating all languages into one training set, and (2) constructing a balanced set with equal representation per language. These setups allow us to examine how data volume and language balance affect cross-lingual generalisation. 

The current study addresses the following research questions (RQs):
\begin{itemize}[noitemsep,topsep=2pt]
     \item[RQ1.] How does the performance of cross-lingual training compare to in-language training for hate-speech detection?

     \item[RQ2.] Would using training data from multiple languages improve performance compared to relying on a single source language?
     
      \item[RQ3.] What could be the potential reasons for the success or failure of transfer learning in hate-speech detection?
\end{itemize}
These questions clarify when cross-lingual transfer is effective for hate-speech detection and identify the conditions under which it succeeds or fails.  \change{Motivated by previous work \cite{bokaei-etal-2025-culture}, we conduct our main experiments with LLaMA-3-8B and additionally replicate them with Gemma-3-4B and Gemma-4-4B to verify that our findings generalize across model families. Our study covers 14 languages: Turkish, Urdu, Persian, Arabic, Hindi, Bengali, Indonesian, Malay, English, Spanish, French, German, Korean, and Chinese.}{}

\section{Related Work}
\textbf{Hate-Speech Detection.}
Early work on hate detection focused primarily on English, applying machine learning and deep learning approaches \cite{asogwa2022hate, davidson2017automated, mullah2021advances, malik2024deep, zimmerman2018improving, 9139953, roy2020framework, zhang2018detecting, bade-etal-2024-social-media, aiyanyo2020systematic, cao2020deephate, risch-etal-2020-offensive, wang-etal-2020-detect, pamungkas-patti-2019-cross, van-hee-etal-2015-detection, guo-gauch-2024-using, cano-basave-etal-2013-weakly}. Later work extended to other languages, including Indonesian \cite{ibrohim-budi-2019-multi}, Danish \cite{sigurbergsson-derczynski-2020-offensive}, Arabic \cite{mubarak-etal-2021-arabic, bensalem2023offensive}, Korean \cite{jeong-etal-2022-kold}, Chinese \cite{deng-etal-2022-cold}, Greek \cite{pitenis-etal-2020-offensive}, Persian \cite{delbari2024spanning}, and Hindi \cite{gupta2022multilingual, kapoor2019mind}.

Recent work highlights key challenges in multilingual hate detection, including comparisons of translation-based and multilingual pipelines \cite{bell2025translate}, limitations in zero-shot transfer for taboo expressions \cite{nozza2021exposing}, and the influence of annotator identity and beliefs on hate judgments \cite{sap2022annotators}. Multimodal datasets such as Multi3Hate reveal substantial cross-cultural variation in hate-speech perception \cite{bui2025multiˆ3hate}, while surveys document strong geographic and cultural imbalances in existing hate-speech resources \cite{tonneau-etal-2024-languages}. Additional studies explore the feasibility of parallel multilingual hate-speech data via machine translation \cite{korre-etal-2024-challenges}. Shared workshops such as OffensEval, HASOC, OSACT5, and GermEval continue to support benchmarking across languages \cite{zampieri-etal-2019-semeval, 10.1145/3368567.3368584, mubarak-etal-2022-overview, wiegand2018overview}.

\textbf{Cross-lingual Transfer Learning for Hate-Speech.}
Transfer learning has been widely applied across NLP tasks, including sentiment analysis \cite{fsih-etal-2022-benchmarking, husain-etal-2022-weak}, irony detection \cite{golazizian-etal-2020-irony}, machine translation \cite{zoph-etal-2016-transfer, kim-etal-2019-pivot, adebara-abdul-mageed-2021-improving}, fake news detection \cite{cruz-etal-2020-localization}, and offensive language classification \cite{zhou-etal-2023-cross}. In multilingual hate detection, prior work studies English-to-low-resource transfer using contextual embeddings \cite{ranasinghe2021multilingual}, compares multilingual and translation-based Arabic--English pipelines \cite{el2022multilingual}, and improves low-resource performance through nearest-neighbor retrieval \cite{ghorbanpour-etal-2025-data}. A recent survey of 67 studies provides a broader overview of cross-lingual offensive language detection methods and datasets \cite{jiang2024cross}.

Work on cultural or social influences on transfer remains limited, though related studies provide evidence. Zero-shot results show that language-specific taboo or colloquial expressions often fail to transfer \cite{nozza2021exposing}. Cross-lingual analyses of English, Chinese, and Korean reveal culture-specific biases affecting transfer, partly mitigated by few-shot adaptation \cite{zhou-etal-2023-cross}. Multimodal work also highlights cross-cultural variation in hate-speech judgments \cite{bui2025multiˆ3hate}. For example, \citet{zhou-etal-2023-cross} find that Korean-to-Chinese transfer can outperform English-to-Chinese, suggesting that contextual or social alignment may support generalisation. Similarly, \citet{bokaei-etal-2025-culture} report stronger transfer into Persian from Arabic and Indonesian than from English. Our study extends this work by analysing transfer across 14 languages and quantifying how dataset, temporal, and typological factors relate to cross-lingual transfer effectiveness.

\begin{table*}[ht]
    \centering
    \scriptsize
    \renewcommand{\arraystretch}{1.2}
    \begin{tabular}{p{1.6cm}p{1.2cm}p{0.9cm}p{0.9cm}p{1.2cm}p{1.2cm}p{1.6cm}p{1.5cm}p{3.4cm}}
        \textbf{Language} & \textbf{Data Size} & \textbf{Train} & \textbf{Test} & \textbf{Train$_{Hate}$} & \textbf{Test$_{Hate}$} & \textbf{Source} & \textbf{Period} & \textbf{Ref.} \\
        \hline
        Bengali (BN) & 30000 & 9007 & 1002 & 4504 & 501 & YT, FB & 2017--2020 & \cite{romim2020hatespeechdetectionbengali} \\
        Urdu (UR) & 10041 & 9000 & 1000 & 4500 & 300 & X & -- & \cite{rizwan-etal-2020-hate} \\
        Hindi (HI) & 8192 & 4171 & 467 & 2222 & 241 & X, FB, WA & -- & \cite{DBLP:journals/corr/abs-2011-03588} \\
        Persian (FA) & 7000 & 2060 & 230 & 1002 & 117 & X & 2020--2022 & \cite{delbari2024spanning} \\
        Indonesian (IN) & 13169 & 9005 & 1003 & 4502 & 502 & X & 2018 & \cite{ibrohim-budi-2019-multi} \\
        Malay (MS) & 5014 & 3403 & 378 & 1701 & 189 & X & 2022--2023 & \cite{maity2023deep} \\
        Arabic (AR) & 15014 & 9005 & 1002 & 4503 & 502 & X & 2016--2018 & \citep{8508247} \\
        Turkish (TR) & 52672 & 9010 & 1003 & 4505 & 502 & X, IG & 2016--2023 & \cite{toraman-etal-2022-large} \\
        Korean (KO) & 9341& 3065 & 300 & 1520 & 142 & X & 2018--2020 & \cite{moon-etal-2020-beep, lee-etal-2022-k} \\
        Chinese (ZH) & 8065 & 4950 & 318 & 2470 & 151 & Web & -- & \cite{moosa2022multilingual} \\
        Spanish (ES) & 29856 & 9000 & 1000 & 4731 & 625 & X & 2016--2023 & \cite{tonneau-etal-2024-languages} \\
        German (DE) & 58025 & 9008 & 1003 & 4503 & 502 & X, FB & 2016--2023 & \cite{tonneau-etal-2024-languages} \\
        French (FR) & 18072 & 7483 & 829 & 3739 & 413 & X & 2016--2023 & \cite{tonneau-etal-2024-languages} \\
        English (EN) & 360494 & 9000 & 1038 & 4500 & 502 & FB, YT, RD, X & 2016--2023 & \cite{tonneau-etal-2024-languages} \\
        \bottomrule
    \end{tabular}
    \caption{Hate language detection datasets. FB = Facebook, YT = YouTube, X = Twitter, WA: WhatsApp, IG = Instagram, RD = Reddit. The definition of Hate and more details for each dataset are shown in Table \ref{tab:hate_definitions}} 
    \label{tab:dataset_details}
\end{table*}

\section{Dataset}
We use publicly available datasets from peer-reviewed publications and shared tasks to ensure quality and comparability. The study covers 14 languages from diverse linguistic families and regions. Most datasets come from Twitter (X), with some including Facebook or YouTube, and collection periods span roughly 2016--2023. Table~\ref{tab:dataset_details} summarises dataset sizes, training splits, hate-sample counts, sources, and collection periods.

All datasets are class-imbalanced, with varying hate/non-hate proportions. For controlled cross-language comparisons, we standardise training size where possible. Among sufficiently large datasets, we use the smallest available hate count, approximately 4,500, and sample an equal number of non-hate instances, yielding balanced datasets of about 10,000 examples per language. We use an 80/10/10 train, validation, and test split. This isolates language effects from dataset volume, preventing high-resource languages such as English or German from appearing stronger due to larger training sets, while keeping computation tractable across 14 models. Balancing was applied only where enough hate examples were available; Hindi, Malay, Korean, Chinese, and Persian fell below this threshold.

To enable consistent evaluation across datasets with fine-grained annotation schemes (e.g., hostile, hate, offensive, abusive), we retain only hate-speech labels and treat all other categories as non-hate. The original label schemes and hate-speech definitions are summarized in Appendix Table~\ref{tab:hate_definitions}. Most datasets adopt a broadly consistent definition of hate-speech as language attacking individuals or groups based on protected identity attributes.

\section{Experimental Setup}

We investigate cross-lingual transfer in hate detection using the datasets summarized in Table \ref{tab:dataset_details}. Our experimental design includes (1) single-language fine-tuning with cross-lingual evaluation, (2) zero-shot baselines, and (3) two complementary multilingual fine-tuning setups to investigate alternative explanations for transfer behavior.

\textbf{Model Selection and Baselines:} \change{We evaluate two types of baselines: (i) zero-shot baselines, including GPT-4o, Gemma-3-4B-Instruct, and LLaMA-3-8B-Instruct; and (ii) fine-tuning baselines, including LLaMA-3-8B, Gemma-3-4B, and Gemma-4-4B. We select LLaMA-3 and the Gemma family because recent benchmarks identify them as strong open-weight choices for multilingual harmful-content detection (\cite{rathore2025fine, bokaei2026benchmarking, raquib2025banhate}), and because they are released under open licenses that permit fine-tuning across all 14 languages on our available hardware. We include two Gemma generations (Gemma-3 and Gemma-4) to test whether transfer structure is stable across model versions as well as across the LLaMA/Gemma architecture divide.
The  fine-tuned baselines outperform all zero-shot baselines across the 14 languages, with LLaMA-3-8B and Gemma-4-4B yielding almost identical average macro-F1 scores, both higher than Gemma-3-4B. We therefore adopt LLaMA-3-8B as our primary model and replicate all main experiments (Solo-Source TL, Global-Imbalanced, Global-Balanced, and Observed Cluster) with Gemma-3-4B, and Gemma-4-4B  to verify that our findings are not specific to one model family. Baseline results are in Table~\ref{tab:multilingual_f1}; full strategy comparisons 
are in Tables~\ref{tab:gemma3_strategies}, and \ref{tab:gemma4_strategies} (Gemma-3 and Gemma-4); full transfer matrices are in Tables~\ref{tab:gemma3_matrix}, and \ref{tab:gemma4_matrix} 
(all in the Appendix). As discussed in Section~\ref{sec:results}, the same qualitative patterns hold across all models, suggesting that the observed transfer structure is not model-specific. A leave-one-language-out analysis on the three largest clusters further confirms the robustness of these gains (Appendix~\ref{app:lolo}).}{}

\begin{figure}
\vspace{-3mm}
\centering
\includegraphics[width=0.82\columnwidth]{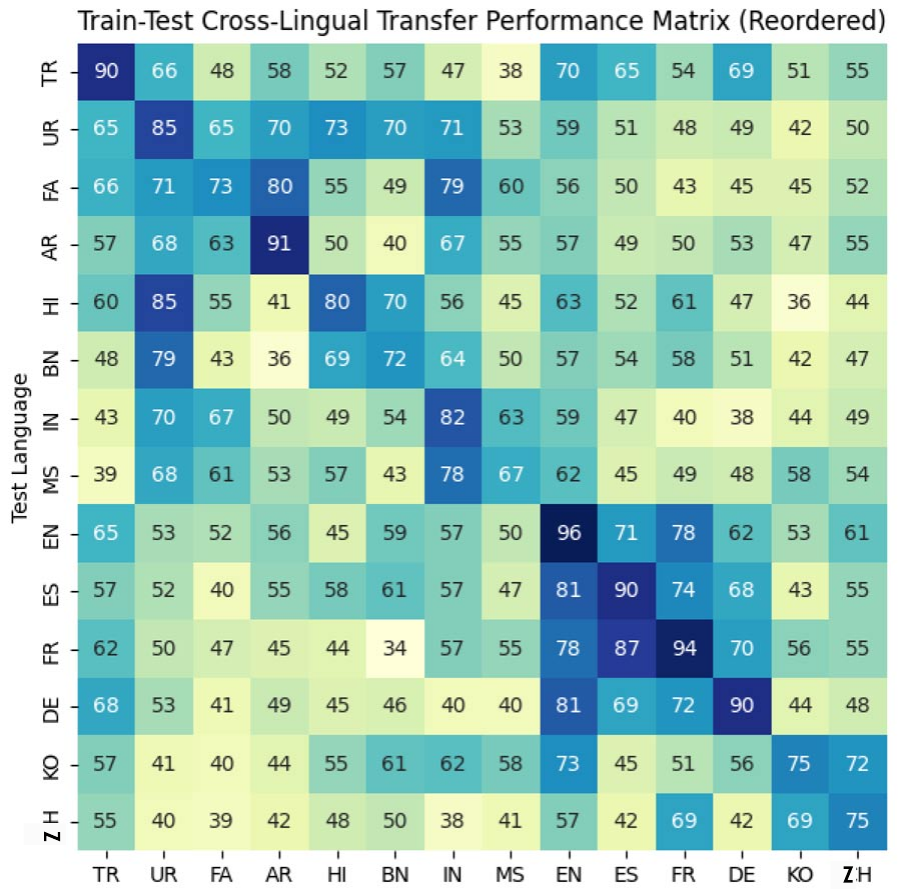}
\vspace{-3mm}
\caption{Cross-lingual F1 scores across 14 languages.}
\vspace{-3mm}
\label{fig:heatmap}
\end{figure}

\textbf{Fine-tuning Configuration:} All fine-tuning uses the official LLaMA-3-8B pretrained checkpoint. We apply AdamW with a learning rate of 1e-5, batch size 8, and a maximum sequence length of 512 tokens. Training proceeds for up to 10 epochs with early stopping (patience = 3) based on validation F1 score. All experiments are run with five random seeds, and we report the mean and standard deviation of F1 across runs to assess result stability. All experiments are conducted on a single NVIDIA Quadro RTX 8000 (48 GB). Performance is reported using macro-F1.

\textbf{Single-Language Fine-Tuning}: For each language, we fine-tune a separate LLaMA-3-8B model on its own training set and evaluate it on all 14 test sets. For example, a model fine-tuned on English is evaluated on English and on the remaining 13 languages. This procedure produces a 14 × 14 cross-lingual transfer matrix, allowing us to examine empirical transfer patterns. Because LLaMA is a multilingual pretrained model, fine-tuning on one language does not erase its multilingual knowledge; rather, it specializes the model toward the hate-speech patterns of the target language.

The 14×14 single-language fine-tuning experiment directly addresses RQ1, revealing which language pairs transfer well or poorly. To address RQ2, we design two complementary multilingual experiments: \textbf{Global Integration (Imbalanced)}, which tests whether greater data volume and multilingual diversity improve generalization, and \textbf{Global Integration (Balanced)}, which removes dataset-size differences as a confounding variable. Together, these setups isolate the effects of data size, multilingual exposure, and the structural similarities surfaced by the 14×14 matrix.

\textbf{ Global Integration (Imbalanced) Experiment}: We merge the full training sets from all 14 languages into one multilingual dataset and fine-tune a single LLaMA-3-8B model. The model is then evaluated separately on each language’s test set. This experiment tests whether greater data volume and diversity produce better generalization.

\textbf{ Global Integration (Balanced) Experiment}: To control for dataset-size differences, we sample an equal number of hate (n = 1000) and non-hate (n = 1000) instances from each language and merge them into a balanced multilingual dataset. A single model is fine-tuned and evaluated on each language’s test set, isolating the effect of balanced representation by controlling for dataset size. 

\change{\textbf{Cluster Tuning:} From the 14×14 transfer matrix, we identify language clusters by inspection, using the criterion that off-diagonal F1 scores are consistently higher within clusters than between cluster members and non-members. The full matrix is reported in Figure~\ref{fig:heatmap} for verification. For each cluster, we merge the training sets and fine-tune one model, then evaluate it separately on each language. To ensure the gains are not driven by a single language, we also conduct leave-one-language-out (LOLO) analysis for the three largest clusters, with full results in Appendix~\ref{app:lolo}.}{} \change{This experiment follows an analyze-then-validate pipeline: empirical structure is first identified from the transfer matrix, then tested through a separate training and evaluation procedure. We note that because the clusters are derived from the same transfer matrix, the cluster-tuning result is confirmatory rather than a prediction on independent data; it tests whether the observed structure is strong enough to yield gains under a distinct training setup, not whether clusters generalize to unseen languages. The LOLO analysis Appendix~\ref{app:lolo}) partially addresses this by showing that each held-out language benefits from a cluster it did not help define. The resulting gains show that observed transfer patterns can inform actionable training strategies, supporting cluster-aware training as a practical approach for multilingual hate-speech detection.}{}

\section{Results}
\label{sec:results}
\setlength{\heavyrulewidth}{0.3pt} 
\setlength{\lightrulewidth}{0.3pt} 

\begin{table*}[!htbp]
\centering
\resizebox{\textwidth}{!}{
\begin{tabular}{lcccccccccccccc}
\toprule
 & TR & UR & FA & AR & HI & BN & IN & MS & EN & ES & FR & DE & KO & ZH \\
\midrule
\small Solo-Source TL 
& 90 \scriptsize$\pm$0.6 & 85 \scriptsize$\pm$0.6 & 80 \scriptsize$\pm$0.8 & 91 \scriptsize$\pm$0.6 & 85 \scriptsize$\pm$0.7 & 79 \scriptsize$\pm$0.7 & 82 \scriptsize$\pm$0.6 & 78 \scriptsize$\pm$0.7 & 96 \scriptsize$\pm$0.5 & 90 \scriptsize$\pm$0.6 & 94 \scriptsize$\pm$0.5 & 90 \scriptsize$\pm$0.6 & \textbf{75} \scriptsize$\pm$0.7 & 75 \scriptsize$\pm$0.7 \\
\midrule

\small Global-Imb 
& \textbf{92} \scriptsize$\pm$0.3 & 88 \scriptsize$\pm$0.3 & 64 \scriptsize$\pm$0.4 & 87 \scriptsize$\pm$0.3 & 73 \scriptsize$\pm$0.4 & 66 \scriptsize$\pm$0.4 & 84 \scriptsize$\pm$0.3 & 62 \scriptsize$\pm$0.4 & \textbf{98} \scriptsize$\pm$0.2 & 95 \scriptsize$\pm$0.3 & \textbf{96} \scriptsize$\pm$0.3 & 94 \scriptsize$\pm$0.3 & 71 \scriptsize$\pm$0.4 & 72 \scriptsize$\pm$0.4 \\
 & {\scriptsize[+2]} & {\scriptsize[+3]} & {\scriptsize[-16]} & {\scriptsize[-4]} & {\scriptsize[-12]} & {\scriptsize[-13]} & {\scriptsize[+2]} & {\scriptsize[-16]} & {\scriptsize[+2]} & {\scriptsize[+5]} & {\scriptsize[+2]} & {\scriptsize[+4]} & {\scriptsize[-4]} & {\scriptsize[-3]} \\

\midrule

\small Global-bal 
& 91 \scriptsize$\pm$0.5 & 87 \scriptsize$\pm$0.5 & 67 \scriptsize$\pm$0.6 & 89 \scriptsize$\pm$0.5 & 81 \scriptsize$\pm$0.6 & 69 \scriptsize$\pm$0.6 & 83 \scriptsize$\pm$0.5 & 63 \scriptsize$\pm$0.6 & 96 \scriptsize$\pm$0.4 & 92 \scriptsize$\pm$0.5 & 94 \scriptsize$\pm$0.4 & 93 \scriptsize$\pm$0.5 & 73 \scriptsize$\pm$0.6 & 76 \scriptsize$\pm$0.6 \\
 & {\scriptsize[+1]} & {\scriptsize[+2]} & {\scriptsize[-13]} & {\scriptsize[-2]} & {\scriptsize[-4]} & {\scriptsize[-10]} & {\scriptsize[+1]} & {\scriptsize[-15]} & {\scriptsize[0]} & {\scriptsize[+2]} & {\scriptsize[0]} & {\scriptsize[+3]} & {\scriptsize[-2]} & {\scriptsize[+1]} \\

\midrule

\small Observed Cluster 
& 90 \scriptsize$\pm$0.4 & \textbf{90} \scriptsize$\pm$0.4 & \textbf{83} \scriptsize$\pm$0.5 & \textbf{93} \scriptsize$\pm$0.4 & \textbf{90} \scriptsize$\pm$0.5 & \textbf{83} \scriptsize$\pm$0.5 & \textbf{85} \scriptsize$\pm$0.4 & \textbf{80} \scriptsize$\pm$0.5 & 97 \scriptsize$\pm$0.3 & \textbf{96} \scriptsize$\pm$0.4 & \textbf{96} \scriptsize$\pm$0.4 & \textbf{95} \scriptsize$\pm$0.4 & \textbf{75} \scriptsize$\pm$0.5 & \textbf{78} \scriptsize$\pm$0.5 \\
 & {\scriptsize[0]} & {\scriptsize[+5]} & {\scriptsize[+3]} & {\scriptsize[+2]} & {\scriptsize[+5]} & {\scriptsize[+4]} & {\scriptsize[+3]} & {\scriptsize[+2]} & {\scriptsize[+1]} & {\scriptsize[+6]} & {\scriptsize[+2]} & {\scriptsize[+5]} & {\scriptsize[0]} & {\scriptsize[+2]} \\

\bottomrule
\end{tabular}
}
\caption{Performance comparison across different fine-tuning strategies. Results are reported as mean $\pm$ standard deviation over five random seeds. Bracketed numbers indicate improvement or decline relative to Solo-Source TL.}
\label{fig:additional}
\end{table*}

Figure \ref{fig:heatmap} presents the 14 × 14 cross-lingual transfer matrix for hate content detection. Each cell reports the F1 score obtained when a model is fine-tuned on the language shown on the x-axis and evaluated on the test language on the y-axis. The matrix contains 196 train–test combinations, providing a comprehensive view of both in-language and cross-lingual performance. Across languages, the highest F1 scores mostly occur on the diagonal, where the model is fine-tuned and tested on the same language. These in-language results are consistently strong: Turkish (90), Arabic (91), English (96), French (94), Spanish (90), and German (90) achieve F1 scores at or above 90. In a few cases, cross-lingual models outperform the in-language baseline; for example, models fine-tuned on Arabic or Indonesian outperform the Persian in-language model on Persian.

Off-diagonal values show substantial variation in the quality of cross-lingual transfer. Based on these off-diagonal patterns, several clusters of languages emerge in which transfer performance is consistently higher within the cluster than across it. These include: (1) Persian–Arabic–Urdu–Indonesian, (2) Urdu–Hindi–Bengali, (3) Indonesian–Malay, (4) Korean–Chinese, and (5) French–Spanish–English–German. As shown in Figure~\ref{fig:heatmap}, models fine-tuned on Urdu perform strongly on Hindi (85), and English-trained models yield relatively high scores on French (78), Spanish (74), and German (69). There are also cases of strong mutual transfer, such as $French \leftrightarrow English$ (78) and $Bengali \leftrightarrow Urdu$ (BN→UR: 70; UR→BN: 79). Lower scores appear for languages like Bengali, whose models perform less well on European test sets (BN→FR: 51; BN→DE: 42), and Chinese models show weaker transfer to several languages (ZH→ES: 42; ZH→AR: 38). Languages such as Korean, Chinese, and Malay generally show lower transfer performance both as train and test languages.

Table \ref{fig:additional} summarizes results from the three complementary experiments. In the Global-Imbalanced setup, combining the full training data from all languages leads to performance gains for several high-resource languages, including English, French, and Spanish (average gain: +2.86), as well as improvements for languages such as Urdu. In contrast, most lower-resource languages—such as Bengali, Persian, and Malay—show declines relative to their single-language baselines (average: -5.43).

In the observed cluster tuning experiment, fine-tuning in language-selective clusters from the 14×14 matrix yields consistent performance improvements for nearly all cluster members (average gain: +5.62). Gains appear across groups such as Hindi-Bengali-Urdu, Arabic-Persian-Indonesian, and English-French-Spanish-German, with varying magnitudes.
In the Global-Balanced setup, equal sampling from all languages reduces the effect of data imbalance. High-resource languages still improve over their single-language baselines, though gains are smaller than in the imbalanced condition (average: +1.57). For lower-resource languages, the balanced dataset mitigates the declines seen under Global-Imbalanced training, producing small gains for most and roughly baseline performance for the remainder.

\change{\textbf{Robustness Across Model Families.} To verify that the observed transfer patterns are not specific to LLaMA-3, we replicate all experimental setups using Gemma-3-4B, and Gemma-4-4B. Although absolute scores differ across models, the same qualitative structure holds for both Gemma-3-4B and Gemma-4-4B. (i) The same five clusters emerge from each model's transfer matrix(Appendix Tables~\ref{tab:gemma3_matrix}--\ref{tab:gemma4_matrix}): for example, Urdu→Hindi, Indonesian→Malay, and the Romance/Germanic group remain among the strongest off-diagonal cells in every model. (ii) Observed Cluster tuning again produces the largest and most consistent gains — average cluster-member gains of +2.4 F1 for Gemma-3-4B (Table 8) and +2.1 for Gemma-4-4B (Table 9), with the same low-resource languages (Persian, Bengali, Malay) benefiting most, mirroring LLaMA-3. (iii) Global-Imbalanced training again favors high-resource languages while harming low-resource ones: Persian, Hindi, Bengali, and Malay drop by 9–14 F1 in both Gemma models, and Global-Balanced training again narrows this gap. The consistency of these three effects across two families and three model generations indicates the transfer structure is a property of the data, not of any single model. Full results are reported in Appendix Tables~\ref{tab:gemma3_strategies}, \ref{tab:gemma4_strategies}, \ref{tab:gemma3_matrix}, and
\ref{tab:gemma4_matrix}.}{}

\section{Analysis}
The 14×14 transfer matrix in Figure~\ref{fig:heatmap} reveals several structural patterns, including the five clusters of languages with strong mutual transfer noted earlier. While most languages achieve their highest performance through in-language learning, certain language pairs also show comparable or better results via transfer. Table~\ref{tab:RQ1} in the Appendix presents hate samples correctly predicted only via transfer learning. We next evaluate each cluster along three dataset-level properties (size, source platform, and temporal overlap) to assess whether these factors account for the observed patterns.

\textbf{Dataset Size:} Across the clusters, dataset size shows partial but inconsistent alignment. The French-Spanish-English-German group comprises datasets with relatively similar hate-sample counts (3.7k-4.7k), which may support more stable within-group transfer. The Urdu–Hindi–Bengali cluster also falls within a relatively similar medium range (3k–5k hate samples), which may support some internal consistency. In contrast, the Persian-Arabic-Urdu-Indonesian cluster spans markedly different dataset sizes: Persian (\textasciitilde1k hate) is much smaller than Arabic, Urdu, or Indonesian (\textasciitilde4.5k each), yet clusters with them. Similarly, Indonesian–Malay shows large variation, as Malay contains substantially fewer hate examples. 

The Korean–Chinese group also exhibits different dataset scales, with Chinese containing roughly twice as many hate examples as Korean. Overall, dataset size aligns with some cluster boundaries (notably the English-French-Spanish-German) but not with clusters involving Persian, Malay, Korean, or Chinese, indicating that size alone cannot explain the structure.

\textbf{Dataset Resource:} Several clusters, including Urdu-Hindi-Bengali, Persian-Arabic-Urdu-Indonesian, and Indonesian-Malay, consist entirely of Twitter-based datasets. However, this does not distinguish them, since most languages in our study, including those outside these clusters, are also Twitter-sourced. Conversely, the Korean-Chinese cluster does not share a single source: Korean is from Twitter, while Chinese comes from general web data. Thus, data source alignment does not systematically explain the cluster structure.

\textbf{Temporal Overlap:} A similar mixed pattern appears when comparing collection periods: the English-French-Spanish--German datasets share nearly identical time spans (2016--2023), which could support more consistent transfer. The Urdu–Hindi–Bengali datasets likewise cover broadly similar years. However, the Persian–Arabic–Urdu–Indonesian cluster shows substantial variation: Persian (2020–22) does not overlap with Arabic (2016–18) or Indonesian ($\le$2018), yet all appear in the same cluster. Similarly, the Indonesian-Malay pair has almost no temporal overlap ($\le$2018 vs.\ 2022--23), yet forms a clear transfer cluster. We cannot evaluate the Korean-Chinese cluster on this dimension, as the collection period for the Chinese dataset is unspecified. These comparisons suggest that temporal alignment may support transfer in some clusters but does not consistently predict cluster structure overall. \change{Table~\ref{tab:cluster_confounds} in Appendix summarizes these dataset-level factors across the five observed clusters.}{}

\subsection{Quantitative Analysis}

We examine three measurable dataset properties: dataset size difference, temporal overlap, and linguistic similarity to assess whether they explain the transfer structure in the matrix. For each of the 182 source--target pairs, we compute size difference as the absolute log-ratio of the hate training sizes, temporal overlap as the intersection-over-union of the datasets' collection-year ranges, and linguistic similarity as the cosine similarity of typological feature vectors from URIEL/lang2vec \cite{bjerva2019language}; pairs with missing collection dates are excluded from the temporal measure. We then compute Spearman correlations \cite{spearman1961proof} between each factor and the off-diagonal F1 scores. As shown in Table~\ref{tab:correlation_results}, linguistic similarity has the strongest association with transfer performance ($\rho = 0.46$), followed by temporal overlap ($\rho = 0.35$), while dataset size difference shows only a weak effect ($\rho = -0.19$). This trend is visible in several typologically close pairs: Urdu$\rightarrow$Hindi attains the highest score in the matrix, and Urdu$\rightarrow$Bengali, English$\rightarrow$German, Spanish$\rightarrow$French, and Indonesian$\rightarrow$Malay are likewise strong. However, these factors alone cannot fully account for the observed patterns: transfer succeeds from Indonesian to Persian (F1=79) and Arabic to Urdu (F1=70) across different language families, while some closely related or temporally overlapping pairs transfer only moderately (e.g., Turkish to Arabic, F1=58). Full derivations and per-pair values are in Appendix~\ref{app:quant_analysis} and Table~\ref{tab:transfer_prop}.

\begin{table}
\centering
\small
\begin{tabular}{lcc}
\toprule
Factor & Spearman $\rho$ & $p$-value \\
\midrule
Dataset size difference & -0.19 & 0.011 \\
Temporal overlap & 0.35 & $1.6 \times 10^{-4}$ \\
Linguistic similarity & 0.46 & $<10^{-9}$ \\
\bottomrule
\end{tabular}
\caption{Spearman correlations between transfer performance and dataset properties.}
\label{tab:correlation_results}
\end{table}

\subsection{Qualitative Analysis} \label{sec:qualitative}
To better understand when transfer occurs, we conduct a cross-lingual qualitative error analysis by inspecting examples correctly classified only after transfer learning and comparing them with those correctly predicted by the in-language model. The patterns we describe below are interpretive observations drawn from manual inspection of representative examples; they are intended to generate hypotheses about why certain clusters transfer well, not to establish cultural similarity as a measured cause.

In the Arabic-Indonesian-Urdu-Persian cluster, Arabic and Urdu particularly support detection of religious and political hate, especially sociopolitical hate in Middle Eastern contexts. Arabic provides cues aligned with Persian discourse, while Urdu contributes strongly to political hate detection, often more than Arabic. Indonesian contributes more to role-based hate, such as profession-targeted hate, and performs well on longer tweets mixing neutral and hateful content.

Table~\ref{tab:RQ1} presents Persian examples correctly predicted only after transfer learning, alongside those captured only by the Persian fine-tuned model. We also observe strong transfer between Indonesian and Malay. Both datasets contain comparable expressions of political frustration, religious commentary, and identity-based accusations. Religious references are frequently used to judge or shame others, including accusations related to religious observance. Similar metaphorical and idiomatic expressions also appear in both languages, including insults referencing moral faults or entities, allowing Indonesian fine-tuned models to generalize well to Malay. Both languages also employ similar proverbs, insults, and metaphorical language, particularly the use of animals such as "\textit{dog}" and "\textit{devil}" as slurs. Social issues such as traffic caused by bazaars, public frustration over commodity scarcity, and emotional sharing on social media are likewise reflected in both contexts. However, the reverse transfer direction is weaker, likely due to the limited amount of hate data in the Malay dataset, which restricts its effectiveness as a source language. A similar pattern appears for Persian, where limited hate training data reduces its effectiveness as a target language. Another cluster where transfer learning outperforms in-language learning includes Urdu, Hindi, and Bengali. Models fine-tuned on Urdu often perform better on Bengali and Hindi than the respective in-language models. Transfer from Urdu is particularly effective for highly polarized, overt hate expressions involving explicit slurs, religious antagonism, or politicized attacks. 

In contrast, in-language models more accurately classify tweets containing idiomatic expressions or contextually grounded criticism. For instance, socially embedded sarcasm in Bengali was frequently missed by the Urdu-fine-tuned model but correctly identified by the Bengali-fine-tuned counterpart. This difference appears related to dataset composition: the Urdu dataset contains a high density of explicit hate markers—religious slurs, profane language, and direct references to identifiable targets—which exhibit lexical consistency and limited idiomatic variation, enabling reliable pattern learning. By comparison, Bengali and Hindi datasets contain more ambiguous, sarcastic, or culturally embedded forms of expression; specifically, the Hindi dataset includes implicit hate, which is harder to generalize across languages. We also observe that transfer strength is not driven by topical similarity alone. Although English and French contain religious or political hate, Arabic or Urdu transfer yields larger gains for detecting such hate in Persian. A similar pattern appears for Arabic-to-English or Arabic-to-French transfer.

To broaden the analysis, we also inspected examples from language pairs outside the same empirical cluster, examining transfer when alignment is weaker. Our results show that even when languages do not belong to the same cluster (e.g., Spanish–Chinese, Persian–French), transfer learning can still detect certain types of hate-speech, though less consistently. In particular, explicit, emotionally extreme, or lexically offensive expressions tend to transfer well across languages. For example, in our experiments, the Spanish→Chinese transfer achieved moderate performance in detecting overtly hateful Chinese instances such as \textit{You’re a brainless feminist bitch} or (\textit{So many feminist bitches on Weibo!}). However, the Spanish-trained model struggled with more context-dependent expressions, such as: \textit{Emperor Tang Minghuang's lifetime reputation was ruined at the hands of the Yang family, A historical reference used to delegitimize women}, or
\textit{I also saw someone called 'Pepsi256' who claims to be a feminist. Lately, she’s been supporting the Sichuan Fine Arts Institute's pro-invasion-of-Taiwan military otakus and happily interacting with the nationalists (a.k.a. little pinks)}.

In contrast, training the model on Korean resulted in significantly better performance not only on overt hate expressions but also on more subtle, context-dependent instances like those discussed above. In other cases, experiments with French and Persian show that a Persian fine-tuned model can still correctly identify highly explicit hate content in French. For example: “Vous pouvez crever sale race”
(You can die, filthy race). This phrase combines an aggressive death wish with a racial slur, making it emotionally extreme and lexically explicit, which enables the Persian fine-tuned model to detect it despite belonging to a different cluster. These results suggest that even across dissimilar clusters, transfer learning succeeds when hate content is lexically explicit or emotionally extreme, but consistently struggles with expressions relying on background social or historical context. 

Taken together, these qualitative observations point to a consistent theme: transfer is strongest for explicit, lexically overt hate and weakest for implicit, culturally or historically grounded expression. We verified these patterns by inspecting the same examples for fine-tuned Gemma-3 and Gemma-4; cases misclassified by LLaMA-3 were typically also misclassified by both Gemma variants, and correctly classified cases showed similar patterns across models, indicating the patterns are not specific to a single model family.

\section{Discussion}
For RQ1, whether cross-lingual training can match in-language hate-speech detection—our results show that cross-lingual models can achieve comparable or superior performance to in-language fine-tuning in several cases. Across the 14 languages studied, this effect is most evident within the observed clusters, where languages such as Persian, Indonesian, Urdu, Malay, Hindi, and Bengali benefit substantially from training on related languages. For RQ2, whether multilingual training improves performance beyond single-source transfer—only specific setups yield consistent gains. Globally imbalanced training primarily benefits high-resource languages while weakening the language-specific signals needed for low-resource ones. In contrast, cluster-based training provides the most reliable improvements across languages.

\change{RQ3 examines potential reasons behind successful or failed transfer. Our quantitative analysis (Table~\ref{tab:correlation_results}) identifies linguistic similarity as the strongest measurable predictor of cross-lingual transfer ($\rho=0.46$), followed by temporal overlap ($\rho=0.35$), while dataset size difference shows only a weak association. These results suggest that transfer patterns in the 14×14 matrix are partly explained by typological and temporal proximity. However, linguistic similarity does not fully account for all patterns: some high-performing pairs, such as Arabic→Urdu, Indonesian→Persian, and Turkish→English, involve typologically different languages, while some closely related pairs show only moderate transfer. The qualitative findings in Section ~\ref{sec:qualitative} suggest that some pairs left unexplained by these measurable factors coincide with shared discourse and communicative patterns within clusters. Because this observation comes from manual inspection, we treat it as a complementary hypothesis alongside linguistic similarity, not as an independently measured factor.}{}

Together, these observations suggest that some of the empirically observed clusters may reflect overlapping cultural cues or communicative patterns \cite{zhou-etal-2023-cross}, rather than transfer arising from any single language-pair property. \change{We caution, however, that this cultural interpretation is complementary to—not a replacement for—the role of linguistic similarity, and that confounds such as topic distribution, annotation scheme, and annotator demographics remain only partly controlled (see Limitations).}{}

Our observations align with prior work on cultural context in hate-speech detection. \citet{lee2023hate} show that models trained in one cultural context often perform poorly across cultures because they miss culturally specific targets and discourse norms. Similarly, \citet{lee2024exploring} show that annotations can vary across countries even within the same language due to cultural and social differences. Our findings extend this work by showing that such differences also affect cross-lingual transfer: shared cultural context may strengthen transfer, while culturally specific expressions can weaken it.

\section {Conclusion}
This study analyzes cross-lingual transfer learning for hate-speech detection across fourteen languages. While in-language fine-tuning remains the most robust baseline, cross-lingual models can match or surpass it by leveraging clusters of languages with similar hate-discourse patterns. Multilingual training is most effective when restricted to these empirical clusters; globally imbalanced integration provides limited support for low-resource languages.\change{Crucially, successful transfer is not fully explained by dataset size, resource availability, or temporal overlap. Among the factors we quantify, linguistic similarity is the strongest predictor; our qualitative analysis additionally points to shared cultural and communicative patterns as a possible but unmeasured contributor within several clusters.
These patterns are consistent across examined models (LLaMA and Gemma), suggesting that the observed transfer structure reflects properties of the data and task. Explicit or highly polarized hate transfers well across distant languages, but culturally embedded and context-dependent forms remain challenging. Overall, effective cross-lingual hate-speech detection benefits from both typological proximity and attention to the social dynamics of online hate. Direct measurement of cross-lingual social alignment, ideally with datasets that hold linguistic similarity constant, is a promising direction for future work.}{}

\section* {Limitations}
This study has several limitations. First, the datasets used across the fourteen languages vary in source platforms, annotation practices, and topical distributions. Although we standardize dataset size where possible and map labels to a binary hate/non-hate scheme, differences in domain, explicitness, and annotation criteria may still influence transfer behavior. In particular, implicit hate is annotated only in a subset of languages, which limits direct evaluation of how implicit expressions transfer. \change{Second, while our quantitative analysis considers dataset size, collection period, and linguistic similarity, several factors that may shape transfer performance remain only partly controlled. These include topic distribution, platform composition, the scope and granularity of annotation schemes, and the demographic characteristics of annotators, which prior work has shown to influence hate-speech judgments \cite{sap2022annotators}. Because these factors may co-vary with the regional or cultural groupings we identify, the empirical clusters we report should be interpreted as observed transfer patterns.}{}

Third, the thematic observations from our qualitative error analysis rely on manual inspection of representative examples and may not capture the full variability of hate discourse across languages; a more extensive evaluation with multiple annotators would offer a stronger foundation for interpreting transferable expressions. \change{Fourth, while our quantitative analysis establishes linguistic similarity as the strongest empirically supported predictor of cross-lingual transfer, the shared cultural and discourse-level patterns identified through qualitative analysis serve as a complementary factor. Direct measurement of cross-lingual sociopolitical alignment was beyond the scope of this work and represents a promising direction for future research, ideally supported by parallel datasets that hold linguistic similarity constant.}{}
Fifth, our cross-model validation spans two families (LLaMA and Gemma) but both models (Gemma-3-4B and Gemma-4-4B) belong to the Gemma line and are smaller than our primary LLaMA-3-8B. The consistency of the transfer structure across LLaMA and Gemma, across two Gemma generations, and across this scale gap is evidence that the patterns are not model-specific, but we cannot rule out that a more distant architecture or a substantially larger model would behave differently. Replication across additional model families and scales is an important direction for future work.

\section*{Ethics Statement}
This study analyzes publicly available hate datasets and does not involve collecting new user data. All datasets were obtained from prior peer-reviewed work or shared tasks that follow established ethical guidelines. Because hate datasets may contain harmful or offensive language, examples shown in this paper are minimized and presented only when necessary for scientific transparency.
No personally identifiable information is included in our datasets or model outputs. All experiments were conducted using anonymized text. Models trained in this work are not intended for deployment without further evaluation, fairness auditing, and context-specific calibration.

\bibliography{ref.bib}

\begin{thebibliography}{75}
\providecommand{\natexlab}[1]{#1}

\bibitem[{Abaskohi et~al.(2024)Abaskohi, Baruni, Masoudi, Abbasi, Babalou,
  Edalat, Kamahi, Mahdizadeh~Sani, Naghavian, Namazifard, Sadeghi, and
  Yaghoobzadeh}]{abaskohi-etal-2024-benchmarking}
Amirhossein Abaskohi, Sara Baruni, Mostafa Masoudi, Nesa Abbasi, Mohammad~Hadi
  Babalou, Ali Edalat, Sepehr Kamahi, Samin Mahdizadeh~Sani, Nikoo Naghavian,
  Danial Namazifard, Pouya Sadeghi, and Yadollah Yaghoobzadeh. 2024.
\newblock \href {https://aclanthology.org/2024.lrec-main.197} {Benchmarking
  large language models for {P}ersian: A preliminary study focusing on
  {C}hat{GPT}}.
\newblock In \emph{Proceedings of the 2024 Joint International Conference on
  Computational Linguistics, Language Resources and Evaluation (LREC-COLING
  2024)}, pages 2189--2203, Torino, Italia. ELRA and ICCL.

\bibitem[{Adebara and Abdul-Mageed(2021)}]{adebara-abdul-mageed-2021-improving}
Ife Adebara and Muhammad Abdul-Mageed. 2021.
\newblock \href {https://aclanthology.org/2021.wmt-1.27/} {Improving similar
  language translation with transfer learning}.
\newblock In \emph{Proceedings of the Sixth Conference on Machine Translation},
  pages 273--278, Online. Association for Computational Linguistics.

\bibitem[{Aiyanyo et~al.(2020)Aiyanyo, Samuel, and Lim}]{aiyanyo2020systematic}
Imatitikua~D Aiyanyo, Hamman Samuel, and Heuiseok Lim. 2020.
\newblock A systematic review of defensive and offensive cybersecurity with
  machine learning.
\newblock \emph{Applied Sciences}, 10(17):5811.

\bibitem[{Albadi et~al.(2018)Albadi, Kurdi, and Mishra}]{8508247}
Nuha Albadi, Maram Kurdi, and Shivakant Mishra. 2018.
\newblock \href {https://doi.org/10.1109/ASONAM.2018.8508247} {Are they our
  brothers? analysis and detection of religious hate speech in the arabic
  twittersphere}.
\newblock In \emph{2018 IEEE/ACM International Conference on Advances in Social
  Networks Analysis and Mining (ASONAM)}, pages 69--76.

\bibitem[{Asogwa et~al.(2022)Asogwa, Chukwuneke, Ngene, and
  Anigbogu}]{asogwa2022hate}
Doris~Chinedu Asogwa, Chiamaka~Ijeoma Chukwuneke, CC~Ngene, and GN~Anigbogu.
  2022.
\newblock Hate speech classification using{SVM} and {N}aive {B}ayes.
\newblock \emph{arXiv preprint arXiv:2204.07057}.

\bibitem[{Bade et~al.(2024)Bade, Kolesnikova, Sidorov, and
  Oropeza}]{bade-etal-2024-social-media}
Girma Bade, Olga Kolesnikova, Grigori Sidorov, and Jos{\'e} Oropeza. 2024.
\newblock \href {https://aclanthology.org/2024.dravidianlangtech-1.40} {Social
  media hate and offensive speech detection using machine learning method}.
\newblock In \emph{Proceedings of the Fourth Workshop on Speech, Vision, and
  Language Technologies for Dravidian Languages}, pages 240--244, St. Julian's,
  Malta. Association for Computational Linguistics.

\bibitem[{Bell et~al.(2025)Bell, S{\'a}nchez, Dale, Stenetorp, Artetxe, and
  Costa-juss{\`a}}]{bell2025translate}
Samuel Bell, Eduardo S{\'a}nchez, David Dale, Pontus Stenetorp, Mikel Artetxe,
  and Marta~R Costa-juss{\`a}. 2025.
\newblock Translate, then detect: Leveraging machine translation for
  cross-lingual toxicity classification.
\newblock In \emph{Proceedings of the Tenth Conference on Machine Translation},
  pages 253--268.

\bibitem[{Bensalem et~al.(2023)Bensalem, Mout, and
  Rosso}]{bensalem2023offensive}
Imene Bensalem, Meryem Mout, and Paolo Rosso. 2023.
\newblock Offensive language detection in {A}rabizi.
\newblock In \emph{Proceedings of ArabicNLP 2023}, pages 423--434.

\bibitem[{Bhardwaj et~al.(2020)Bhardwaj, Akhtar, Ekbal, Das, and
  Chakraborty}]{DBLP:journals/corr/abs-2011-03588}
Mohit Bhardwaj, Md.~Shad Akhtar, Asif Ekbal, Amitava Das, and Tanmoy
  Chakraborty. 2020.
\newblock \href {https://arxiv.org/abs/2011.03588} {Hostility detection dataset
  in hindi}.
\newblock \emph{CoRR}, abs/2011.03588.

\bibitem[{Bjerva et~al.(2019)Bjerva, {\"O}stling, Veiga, Tiedemann, and
  Augenstein}]{bjerva2019language}
Johannes Bjerva, Robert {\"O}stling, Maria~Han Veiga, J{\"o}rg Tiedemann, and
  Isabelle Augenstein. 2019.
\newblock What do language representations really represent?
\newblock \emph{Computational Linguistics}, 45(2):381--389.

\bibitem[{Bokaei et~al.(2025)Bokaei, Magdy, and
  Webber}]{bokaei-etal-2025-culture}
Zahra Bokaei, Walid Magdy, and Bonnie Webber. 2025.
\newblock \href {https://doi.org/10.18653/v1/2025.acl-long.456} {Culture
  matters in toxic language detection in {P}ersian}.
\newblock In \emph{Proceedings of the 63rd Annual Meeting of the Association
  for Computational Linguistics (Volume 1: Long Papers)}, pages 9290--9304,
  Vienna, Austria. Association for Computational Linguistics.

\bibitem[{Bokaei et~al.(2026{\natexlab{a}})Bokaei, Magdy, and
  Webber}]{BOKAEI2026104912}
Zahra Bokaei, Walid Magdy, and Bonnie Webber. 2026{\natexlab{a}}.
\newblock \href {https://doi.org/10.1016/j.ipm.2026.104912} {Farsi natural
  language processing: A survey}.
\newblock \emph{Information Processing \& Management}, 63(8):104912.

\bibitem[{Bokaei et~al.(2026{\natexlab{b}})Bokaei, Webber, and
  Magdy}]{bokaei2026benchmarking}
Zahra Bokaei, Bonnie Webber, and Walid Magdy. 2026{\natexlab{b}}.
\newblock Benchmarking offensive language detection in persian and pashto.
\newblock In \emph{The Proceedings of the First Workshop on NLP and LLMs for
  the Iranian Language Family}, pages 13--23.

\bibitem[{Bui et~al.(2025)Bui, Von Der~Wense, and
  Lauscher}]{bui2025multiˆ3hate}
Minh~Duc Bui, Katharina Von Der~Wense, and Anne Lauscher. 2025.
\newblock Multiˆ3hate: Multimodal, multilingual, and multicultural hate speech
  detection with vision--language models.
\newblock In \emph{Proceedings of the 2025 Conference of the Nations of the
  Americas Chapter of the Association for Computational Linguistics: Human
  Language Technologies (Volume 1: Long Papers)}, pages 9714--9731.

\bibitem[{Cano~Basave et~al.(2013)Cano~Basave, He, Liu, and
  Zhao}]{cano-basave-etal-2013-weakly}
Amparo~Elizabeth Cano~Basave, Yulan He, Kang Liu, and Jun Zhao. 2013.
\newblock \href {https://aclanthology.org/I13-1013} {A weakly supervised
  {B}ayesian model for violence detection in social media}.
\newblock In \emph{Proceedings of the Sixth International Joint Conference on
  Natural Language Processing}, pages 109--117, Nagoya, Japan. Asian Federation
  of Natural Language Processing.

\bibitem[{Cao et~al.(2020)Cao, Lee, and Hoang}]{cao2020deephate}
Rui Cao, Roy Ka-Wei Lee, and Tuan-Anh Hoang. 2020.
\newblock Deephate: Hate speech detection via multi-faceted text
  representations.
\newblock In \emph{Proceedings of the 12th ACM Conference on Web Science},
  pages 11--20.

\bibitem[{Cruz et~al.(2020)Cruz, Tan, and Cheng}]{cruz-etal-2020-localization}
Jan Christian~Blaise Cruz, Julianne~Agatha Tan, and Charibeth Cheng. 2020.
\newblock \href {https://aclanthology.org/2020.lrec-1.316/} {Localization of
  fake news detection via multitask transfer learning}.
\newblock In \emph{Proceedings of the Twelfth Language Resources and Evaluation
  Conference}, pages 2596--2604, Marseille, France. European Language Resources
  Association.

\bibitem[{Davidson et~al.(2017)Davidson, Warmsley, Macy, and
  Weber}]{davidson2017automated}
Thomas Davidson, Dana Warmsley, Michael Macy, and Ingmar Weber. 2017.
\newblock Automated hate speech detection and the problem of offensive
  language.
\newblock In \emph{Proceedings of the international AAAI conference on web and
  social media}, volume~11, pages 512--515.

\bibitem[{Delbari et~al.(2024)Delbari, Moosavi, and
  Pilehvar}]{delbari2024spanning}
Zahra Delbari, Nafise~Sadat Moosavi, and Mohammad~Taher Pilehvar. 2024.
\newblock Spanning the spectrum of hatred detection: a {P}ersian multi-label
  hate speech dataset with annotator rationales.
\newblock In \emph{Proceedings of the AAAI Conference on Artificial
  Intelligence}, volume~38, pages 17889--17897.

\bibitem[{Deng et~al.(2022)Deng, Zhou, Sun, Zheng, Mi, Meng, and
  Huang}]{deng-etal-2022-cold}
Jiawen Deng, Jingyan Zhou, Hao Sun, Chujie Zheng, Fei Mi, Helen Meng, and
  Minlie Huang. 2022.
\newblock \href {https://doi.org/10.18653/v1/2022.emnlp-main.796} {{COLD}: A
  benchmark for {C}hinese offensive language detection}.
\newblock In \emph{Proceedings of the 2022 Conference on Empirical Methods in
  Natural Language Processing}, pages 11580--11599, Abu Dhabi, United Arab
  Emirates. Association for Computational Linguistics.

\bibitem[{El-Alami et~al.(2022)El-Alami, El~Alaoui, and
  Nahnahi}]{el2022multilingual}
Fatima-zahra El-Alami, Said~Ouatik El~Alaoui, and Noureddine~En Nahnahi. 2022.
\newblock A multilingual offensive language detection method based on transfer
  learning from transformer fine-tuning model.
\newblock \emph{Journal of King Saud University-Computer and Information
  Sciences}, 34(8):6048--6056.

\bibitem[{Fsih et~al.(2022)Fsih, Kchaou, Boujelbane, and
  Hadrich-Belguith}]{fsih-etal-2022-benchmarking}
Emna Fsih, Sameh Kchaou, Rahma Boujelbane, and Lamia Hadrich-Belguith. 2022.
\newblock \href {https://doi.org/10.18653/v1/2022.wanlp-1.44} {Benchmarking
  transfer learning approaches for sentiment analysis of {A}rabic dialect}.
\newblock In \emph{Proceedings of the Seventh Arabic Natural Language
  Processing Workshop (WANLP)}, pages 431--435, Abu Dhabi, United Arab Emirates
  (Hybrid). Association for Computational Linguistics.

\bibitem[{Ghorbanpour et~al.(2025)Ghorbanpour, Dementieva, and
  Fraser}]{ghorbanpour-etal-2025-data}
Faeze Ghorbanpour, Daryna Dementieva, and Alexander Fraser. 2025.
\newblock \href {https://doi.org/10.18653/v1/2025.emnlp-main.1507}
  {Data-efficient hate speech detection via cross-lingual nearest neighbor
  retrieval with limited labeled data}.
\newblock In \emph{Proceedings of the 2025 Conference on Empirical Methods in
  Natural Language Processing}, pages 29662--29680, Suzhou, China. Association
  for Computational Linguistics.

\bibitem[{Golazizian et~al.(2020)Golazizian, Sabeti, Ashrafi~Asli, Majdabadi,
  Momenzadeh, and Fahmi}]{golazizian-etal-2020-irony}
Preni Golazizian, Behnam Sabeti, Seyed~Arad Ashrafi~Asli, Zahra Majdabadi, Omid
  Momenzadeh, and Reza Fahmi. 2020.
\newblock \href {https://aclanthology.org/2020.lrec-1.346/} {Irony detection in
  {P}ersian language: A transfer learning approach using emoji prediction}.
\newblock In \emph{Proceedings of the Twelfth Language Resources and Evaluation
  Conference}, pages 2839--2845, Marseille, France. European Language Resources
  Association.

\bibitem[{Guo and Gauch(2024)}]{guo-gauch-2024-using}
Xiaoyu Guo and Susan Gauch. 2024.
\newblock \href {https://aclanthology.org/2024.trac-1.7} {Using sarcasm to
  improve cyberbullying detection}.
\newblock In \emph{Proceedings of the Fourth Workshop on Threat, Aggression
  {\&} Cyberbullying @ LREC-COLING-2024}, pages 52--59, Torino, Italia. ELRA
  and ICCL.

\bibitem[{Gupta et~al.(2022)Gupta, Roychowdhury, Das, Banerjee, Saha, Mathew,
  Mukherjee et~al.}]{gupta2022multilingual}
Vikram Gupta, Sumegh Roychowdhury, Mithun Das, Somnath Banerjee, Punyajoy Saha,
  Binny Mathew, Animesh Mukherjee, et~al. 2022.
\newblock Multilingual abusive comment detection at scale for {I}ndic
  languages.
\newblock \emph{Advances in Neural Information Processing Systems},
  35:26176--26191.

\bibitem[{Hoang et~al.(2024)Hoang, Do, Do, Vu, and Luu}]{hoang-etal-2024-toxcl}
Nhat Hoang, Xuan~Long Do, Duc~Anh Do, Duc~Anh Vu, and Anh~Tuan Luu. 2024.
\newblock \href {https://doi.org/10.18653/v1/2024.naacl-long.359} {{T}o{XCL}: A
  unified framework for toxic speech detection and explanation}.
\newblock In \emph{Proceedings of the 2024 Conference of the North American
  Chapter of the Association for Computational Linguistics: Human Language
  Technologies (Volume 1: Long Papers)}, pages 6460--6472, Mexico City, Mexico.
  Association for Computational Linguistics.

\bibitem[{Husain et~al.(2022)Husain, Al-Ostad, and
  Omar}]{husain-etal-2022-weak}
Fatemah Husain, Hana Al-Ostad, and Halima Omar. 2022.
\newblock \href {https://doi.org/10.18653/v1/2022.wanlp-1.15} {A weak
  supervised transfer learning approach for sentiment analysis to the {K}uwaiti
  dialect}.
\newblock In \emph{Proceedings of the Seventh Arabic Natural Language
  Processing Workshop (WANLP)}, pages 161--173, Abu Dhabi, United Arab Emirates
  (Hybrid). Association for Computational Linguistics.

\bibitem[{Ibrohim and Budi(2019)}]{ibrohim-budi-2019-multi}
Muhammad~Okky Ibrohim and Indra Budi. 2019.
\newblock \href {https://doi.org/10.18653/v1/W19-3506} {Multi-label hate speech
  and abusive language detection in {I}ndonesian {T}witter}.
\newblock In \emph{Proceedings of the Third Workshop on Abusive Language
  Online}, pages 46--57, Florence, Italy. Association for Computational
  Linguistics.

\bibitem[{Jeong et~al.(2022)Jeong, Oh, Lee, Ahn, Moon, Park, and
  Oh}]{jeong-etal-2022-kold}
Younghoon Jeong, Juhyun Oh, Jongwon Lee, Jaimeen Ahn, Jihyung Moon, Sungjoon
  Park, and Alice Oh. 2022.
\newblock \href {https://doi.org/10.18653/v1/2022.emnlp-main.744} {{KOLD}:
  {K}orean offensive language dataset}.
\newblock In \emph{Proceedings of the 2022 Conference on Empirical Methods in
  Natural Language Processing}, pages 10818--10833, Abu Dhabi, United Arab
  Emirates. Association for Computational Linguistics.

\bibitem[{Jiang and Zubiaga(2024)}]{jiang2024cross}
Aiqi Jiang and Arkaitz Zubiaga. 2024.
\newblock Cross-lingual offensive language detection: A systematic review of
  datasets, transfer approaches and challenges.
\newblock \emph{arXiv preprint arXiv:2401.09244}.

\bibitem[{Kapoor et~al.(2019)Kapoor, Kumar, Rajput, Shah, Kumaraguru, and
  Zimmermann}]{kapoor2019mind}
Raghav Kapoor, Yaman Kumar, Kshitij Rajput, Rajiv~Ratn Shah, Ponnurangam
  Kumaraguru, and Roger Zimmermann. 2019.
\newblock Mind your language: Abuse and offense detection for code-switched
  languages.
\newblock In \emph{Proceedings of the AAAI conference on artificial
  intelligence}, volume~33, pages 9951--9952.

\bibitem[{Khondaker et~al.(2023)Khondaker, Waheed, Nagoudi, and
  Abdul-Mageed}]{khondaker-etal-2023-gptaraeval}
Md~Tawkat~Islam Khondaker, Abdul Waheed, El~Moatez~Billah Nagoudi, and Muhammad
  Abdul-Mageed. 2023.
\newblock \href {https://doi.org/10.18653/v1/2023.emnlp-main.16}
  {{GPTA}ra{E}val: A comprehensive evaluation of {C}hat{GPT} on {A}rabic
  {NLP}}.
\newblock In \emph{Proceedings of the 2023 Conference on Empirical Methods in
  Natural Language Processing}, pages 220--247, Singapore. Association for
  Computational Linguistics.

\bibitem[{Kim et~al.(2019)Kim, Petrov, Petrushkov, Khadivi, and
  Ney}]{kim-etal-2019-pivot}
Yunsu Kim, Petre Petrov, Pavel Petrushkov, Shahram Khadivi, and Hermann Ney.
  2019.
\newblock \href {https://doi.org/10.18653/v1/D19-1080} {Pivot-based transfer
  learning for neural machine translation between non-{E}nglish languages}.
\newblock In \emph{Proceedings of the 2019 Conference on Empirical Methods in
  Natural Language Processing and the 9th International Joint Conference on
  Natural Language Processing (EMNLP-IJCNLP)}, pages 866--876, Hong Kong,
  China. Association for Computational Linguistics.

\bibitem[{Korre et~al.(2024)Korre, Muti, and
  Barr{\'o}n-Cede{\~n}o}]{korre-etal-2024-challenges}
Katerina Korre, Arianna Muti, and Alberto Barr{\'o}n-Cede{\~n}o. 2024.
\newblock \href {https://aclanthology.org/2024.lrec-main.1376/} {The challenges
  of creating a parallel multilingual hate speech corpus: An exploration}.
\newblock In \emph{Proceedings of the 2024 Joint International Conference on
  Computational Linguistics, Language Resources and Evaluation (LREC-COLING
  2024)}, pages 15842--15853, Torino, Italia. ELRA and ICCL.

\bibitem[{Kumar et~al.(2024)Kumar, Sharma, and Bedi}]{kumar2024towards}
Ankit Kumar, Richa Sharma, and Punam Bedi. 2024.
\newblock Towards optimal {NLP} solutions: Analyzing {GPT} and {LLaMA}-2 models
  across model scale, dataset size, and task diversity.
\newblock \emph{Engineering, Technology \& Applied Science Research},
  14(3):14219--14224.

\bibitem[{Lee et~al.(2022)Lee, Lim, Lee, Jo, Kim, Yoon, and
  Han}]{lee-etal-2022-k}
Jean Lee, Taejun Lim, Heejun Lee, Bogeun Jo, Yangsok Kim, Heegeun Yoon, and
  Soyeon~Caren Han. 2022.
\newblock \href {https://aclanthology.org/2022.coling-1.311/} {K-{MH}a{S}: A
  multi-label hate speech detection dataset in {K}orean online news comment}.
\newblock In \emph{Proceedings of the 29th International Conference on
  Computational Linguistics}, pages 3530--3538, Gyeongju, Republic of Korea.
  International Committee on Computational Linguistics.

\bibitem[{Lee et~al.(2024)Lee, Jung, Myung, Jin, Camacho-Collados, Kim, and
  Oh}]{lee2024exploring}
Nayeon Lee, Chani Jung, Junho Myung, Jiho Jin, Jose Camacho-Collados, Juho Kim,
  and Alice Oh. 2024.
\newblock Exploring cross-cultural differences in english hate speech
  annotations: From dataset construction to analysis.
\newblock In \emph{Proceedings of the 2024 Conference of the North American
  Chapter of the Association for Computational Linguistics: Human Language
  Technologies (Volume 1: Long Papers)}, pages 4205--4224.

\bibitem[{Lee et~al.(2023)Lee, Jung, and Oh}]{lee2023hate}
Nayeon Lee, Chani Jung, and Alice Oh. 2023.
\newblock Hate speech classifiers are culturally insensitive.
\newblock In \emph{Proceedings of the first workshop on cross-cultural
  considerations in NLP (C3NLP)}, pages 35--46.

\bibitem[{Maity et~al.(2023)Maity, Bhattacharya, Saha, and
  Seera}]{maity2023deep}
Krishanu Maity, Shaubhik Bhattacharya, Sriparna Saha, and Manjeevan Seera.
  2023.
\newblock A deep learning framework for the detection of malay hate speech.
\newblock \emph{IEEE Access}, 11:79542--79552.

\bibitem[{Malik et~al.(2024)Malik, Qiao, Pang, and van~den
  Hengel}]{malik2024deep}
Jitendra~Singh Malik, Hezhe Qiao, Guansong Pang, and Anton van~den Hengel.
  2024.
\newblock Deep learning for hate speech detection: a comparative study.
\newblock \emph{International Journal of Data Science and Analytics}, pages
  1--16.

\bibitem[{Mandl et~al.(2019)Mandl, Modha, Majumder, Patel, Dave, Mandlia, and
  Patel}]{10.1145/3368567.3368584}
Thomas Mandl, Sandip Modha, Prasenjit Majumder, Daksh Patel, Mohana Dave,
  Chintak Mandlia, and Aditya Patel. 2019.
\newblock \href {https://doi.org/10.1145/3368567.3368584} {Overview of the
  hasoc track at fire 2019: Hate speech and offensive content identification in
  {I}ndo-{E}uropean languages}.
\newblock In \emph{Proceedings of the 11th Annual Meeting of the Forum for
  Information Retrieval Evaluation}, FIRE '19, page 14–17, New York, NY, USA.
  Association for Computing Machinery.

\bibitem[{Meyer and Buys(2024)}]{meyer-buys-2024-systematic}
Francois Meyer and Jan Buys. 2024.
\newblock \href {https://doi.org/10.18653/v1/2024.findings-naacl.141} {A
  systematic analysis of subwords and cross-lingual transfer in multilingual
  translation}.
\newblock In \emph{Findings of the Association for Computational Linguistics:
  NAACL 2024}, pages 2194--2200, Mexico City, Mexico. Association for
  Computational Linguistics.

\bibitem[{Moon et~al.(2020)Moon, Cho, and Lee}]{moon-etal-2020-beep}
Jihyung Moon, Won~Ik Cho, and Junbum Lee. 2020.
\newblock \href {https://doi.org/10.18653/v1/2020.socialnlp-1.4} {{BEEP}!
  {K}orean corpus of online news comments for toxic speech detection}.
\newblock In \emph{Proceedings of the Eighth International Workshop on Natural
  Language Processing for Social Media}, pages 25--31, Online. Association for
  Computational Linguistics.

\bibitem[{Moosa and Najiba(2022)}]{moosa2022multilingual}
Wajid~Hassan Moosa and Najiba. 2022.
\newblock Multilingual hate speech dataset.
\newblock
  \url{https://www.kaggle.com/datasets/wajidhassanmoosa/multilingual-hatespeech-dataset}.
\newblock Open Data Commons Attribution License (ODC-By) v1.0.

\bibitem[{Mubarak et~al.(2022)Mubarak, Al-Khalifa, and
  Al-Thubaity}]{mubarak-etal-2022-overview}
Hamdy Mubarak, Hend Al-Khalifa, and Abdulmohsen Al-Thubaity. 2022.
\newblock \href {https://aclanthology.org/2022.osact-1.20} {Overview of
  {OSACT}5 shared task on {A}rabic offensive language and hate speech
  detection}.
\newblock In \emph{Proceedinsg of the 5th Workshop on Open-Source Arabic
  Corpora and Processing Tools with Shared Tasks on Qur'an QA and Fine-Grained
  Hate Speech Detection}, pages 162--166, Marseille, France. European Language
  Resources Association.

\bibitem[{Mubarak et~al.(2021)Mubarak, Rashed, Darwish, Samih, and
  Abdelali}]{mubarak-etal-2021-arabic}
Hamdy Mubarak, Ammar Rashed, Kareem Darwish, Younes Samih, and Ahmed Abdelali.
  2021.
\newblock \href {https://aclanthology.org/2021.wanlp-1.13} {{A}rabic offensive
  language on {T}witter: Analysis and experiments}.
\newblock In \emph{Proceedings of the Sixth Arabic Natural Language Processing
  Workshop}, pages 126--135, Kyiv, Ukraine (Virtual). Association for
  Computational Linguistics.

\bibitem[{Mullah and Zainon(2021)}]{mullah2021advances}
Nanlir~Sallau Mullah and Wan Mohd Nazmee~Wan Zainon. 2021.
\newblock Advances in machine learning algorithms for hate speech detection in
  social media: a review.
\newblock \emph{IEEE Access}, 9:88364--88376.

\bibitem[{Nozza(2021)}]{nozza2021exposing}
Debora Nozza. 2021.
\newblock Exposing the limits of zero-shot cross-lingual hate speech detection.
\newblock In \emph{Proceedings of the 59th Annual Meeting of the Association
  for Computational Linguistics and the 11th International Joint Conference on
  Natural Language Processing (Volume 2: Short Papers)}, pages 907--914.

\bibitem[{Pamungkas and Patti(2019)}]{pamungkas-patti-2019-cross}
Endang~Wahyu Pamungkas and Viviana Patti. 2019.
\newblock \href {https://doi.org/10.18653/v1/P19-2051} {Cross-domain and
  cross-lingual abusive language detection: A hybrid approach with deep
  learning and a multilingual lexicon}.
\newblock In \emph{Proceedings of the 57th Annual Meeting of the Association
  for Computational Linguistics: Student Research Workshop}, pages 363--370,
  Florence, Italy. Association for Computational Linguistics.

\bibitem[{Pawar et~al.(2024)Pawar, Park, Jin, Haznitrama, Myung, Oh, Augenstein
  et~al.}]{pawar2024cultural}
Siddhesh Pawar, Junyeong Park, Jiho Jin, Faiz~Ghifari Haznitrama, Junho Myung,
  Alice Oh, Isabelle Augenstein, et~al. 2024.
\newblock Survey of cultural awareness in language models: Text and beyond.
\newblock \emph{arXiv preprint arXiv:2411.00860}.

\bibitem[{Pitenis et~al.(2020)Pitenis, Zampieri, and
  Ranasinghe}]{pitenis-etal-2020-offensive}
Zesis Pitenis, Marcos Zampieri, and Tharindu Ranasinghe. 2020.
\newblock \href {https://aclanthology.org/2020.lrec-1.629} {Offensive language
  identification in {G}reek}.
\newblock In \emph{Proceedings of the Twelfth Language Resources and Evaluation
  Conference}, pages 5113--5119, Marseille, France. European Language Resources
  Association.

\bibitem[{Ranasinghe and Zampieri(2021)}]{ranasinghe2021multilingual}
Tharindu Ranasinghe and Marcos Zampieri. 2021.
\newblock Multilingual offensive language identification for low-resource
  languages.
\newblock \emph{Transactions on Asian and Low-Resource Language Information
  Processing}, 21(1):1--13.

\bibitem[{Raquib et~al.(2025)Raquib, Mazumder, Fuad, Ishmam, and
  Fahim}]{raquib2025banhate}
Faisal~Hossain Raquib, Akm Moshiur~Rahman Mazumder, Md~Tahmid~Hasan Fuad,
  Md~Farhan Ishmam, and Md~Fahim. 2025.
\newblock Banhate: An up-to-date and fine-grained bangla hate speech dataset.
\newblock In \emph{Proceedings of the Second Workshop on Bangla Language
  Processing (BLP-2025)}, pages 237--248.

\bibitem[{Rathore and Chaurasia(2025)}]{rathore2025fine}
Bhawani~Singh Rathore and Sandeep Chaurasia. 2025.
\newblock Fine tuning large language models for hate speech detection in
  hinglish and code mixed custom dataset through a socially responsible
  approach for safer digital platforms.
\newblock \emph{Discover Sustainability}, 6(1):1409.

\bibitem[{Risch et~al.(2020)Risch, Ruff, and
  Krestel}]{risch-etal-2020-offensive}
Julian Risch, Robin Ruff, and Ralf Krestel. 2020.
\newblock \href {https://aclanthology.org/2020.trac-1.22} {Offensive language
  detection explained}.
\newblock In \emph{Proceedings of the Second Workshop on Trolling, Aggression
  and Cyberbullying}, pages 137--143, Marseille, France. European Language
  Resources Association (ELRA).

\bibitem[{Rizwan et~al.(2020)Rizwan, Shakeel, and
  Karim}]{rizwan-etal-2020-hate}
Hammad Rizwan, Muhammad~Haroon Shakeel, and Asim Karim. 2020.
\newblock \href {https://doi.org/10.18653/v1/2020.emnlp-main.197} {Hate-speech
  and offensive language detection in {R}oman {U}rdu}.
\newblock In \emph{Proceedings of the 2020 Conference on Empirical Methods in
  Natural Language Processing (EMNLP)}, pages 2512--2522, Online. Association
  for Computational Linguistics.

\bibitem[{Romim et~al.(2020)Romim, Ahmed, Talukder, and
  Islam}]{romim2020hatespeechdetectionbengali}
Nauros Romim, Mosahed Ahmed, Hriteshwar Talukder, and Md~Saiful Islam. 2020.
\newblock \href {https://arxiv.org/abs/2012.09686} {Hate speech detection in
  the bengali language: A dataset and its baseline evaluation}.
\newblock \emph{Preprint}, arXiv:2012.09686.

\bibitem[{Roy et~al.(2020)Roy, Tripathy, Das, and Gao}]{roy2020framework}
Pradeep~Kumar Roy, Asis~Kumar Tripathy, Tapan~Kumar Das, and Xiao-Zhi Gao.
  2020.
\newblock A framework for hate speech detection using deep convolutional neural
  network.
\newblock \emph{IEEE Access}, 8:204951--204962.

\bibitem[{Sap et~al.(2022{\natexlab{a}})Sap, Swayamdipta, Vianna, Zhou, Choi,
  and Smith}]{sap-etal-2022-annotators}
Maarten Sap, Swabha Swayamdipta, Laura Vianna, Xuhui Zhou, Yejin Choi, and
  Noah~A. Smith. 2022{\natexlab{a}}.
\newblock \href {https://doi.org/10.18653/v1/2022.naacl-main.431} {Annotators
  with attitudes: How annotator beliefs and identities bias toxic language
  detection}.
\newblock In \emph{Proceedings of the 2022 Conference of the North American
  Chapter of the Association for Computational Linguistics: Human Language
  Technologies}, pages 5884--5906, Seattle, United States. Association for
  Computational Linguistics.

\bibitem[{Sap et~al.(2022{\natexlab{b}})Sap, Swayamdipta, Vianna, Zhou, Choi,
  and Smith}]{sap2022annotators}
Maarten Sap, Swabha Swayamdipta, Laura Vianna, Xuhui Zhou, Yejin Choi, and
  Noah~A Smith. 2022{\natexlab{b}}.
\newblock Annotators with attitudes: How annotator beliefs and identities bias
  toxic language detection.
\newblock In \emph{Proceedings of the 2022 conference of the north american
  chapter of the association for computational linguistics: Human language
  technologies}, pages 5884--5906.

\bibitem[{Sigurbergsson and
  Derczynski(2020)}]{sigurbergsson-derczynski-2020-offensive}
Gudbjartur~Ingi Sigurbergsson and Leon Derczynski. 2020.
\newblock \href {https://aclanthology.org/2020.lrec-1.430} {Offensive language
  and hate speech detection for {D}anish}.
\newblock In \emph{Proceedings of the Twelfth Language Resources and Evaluation
  Conference}, pages 3498--3508, Marseille, France. European Language Resources
  Association.

\bibitem[{Spearman(1961)}]{spearman1961proof}
Charles Spearman. 1961.
\newblock The proof and measurement of association between two things.

\bibitem[{Tonneau(2024)}]{tonneau2024spanish}
Manuel Tonneau. 2024.
\newblock \href
  {https://huggingface.co/datasets/manueltonneau/spanish-hate-speech-superset}
  {Spanish hate speech superset}.
\newblock Accessed: 2025-05-17.

\bibitem[{Tonneau et~al.(2024)Tonneau, Liu, Fraiberger, Schroeder, Hale, and
  R{\"o}ttger}]{tonneau-etal-2024-languages}
Manuel Tonneau, Diyi Liu, Samuel Fraiberger, Ralph Schroeder, Scott Hale, and
  Paul R{\"o}ttger. 2024.
\newblock \href {https://aclanthology.org/2024.emnlp-main.123} {From languages
  to geographies: Towards evaluating cultural bias in hate speech datasets}.
\newblock In \emph{Proceedings of the 2024 Conference on Empirical Methods in
  Natural Language Processing (EMNLP)}.

\bibitem[{Toraman et~al.(2022)Toraman, {\c{S}}ahinu{\c{c}}, and
  Yilmaz}]{toraman-etal-2022-large}
Cagri Toraman, Furkan {\c{S}}ahinu{\c{c}}, and Eyup Yilmaz. 2022.
\newblock \href {https://aclanthology.org/2022.lrec-1.238/} {Large-scale hate
  speech detection with cross-domain transfer}.
\newblock In \emph{Proceedings of the Thirteenth Language Resources and
  Evaluation Conference}, pages 2215--2225, Marseille, France. European
  Language Resources Association.

\bibitem[{Van~Hee et~al.(2015)Van~Hee, Lefever, Verhoeven, Mennes, Desmet,
  De~Pauw, Daelemans, and Hoste}]{van-hee-etal-2015-detection}
Cynthia Van~Hee, Els Lefever, Ben Verhoeven, Julie Mennes, Bart Desmet, Guy
  De~Pauw, Walter Daelemans, and Veronique Hoste. 2015.
\newblock \href {https://aclanthology.org/R15-1086} {Detection and fine-grained
  classification of cyberbullying events}.
\newblock In \emph{Proceedings of the International Conference Recent Advances
  in Natural Language Processing}, pages 672--680, Hissar, Bulgaria. INCOMA
  Ltd. Shoumen, BULGARIA.

\bibitem[{Wang et~al.(2020)Wang, Lu, Han, Long, and
  Poon}]{wang-etal-2020-detect}
Kunze Wang, Dong Lu, Caren Han, Siqu Long, and Josiah Poon. 2020.
\newblock \href {https://doi.org/10.18653/v1/2020.coling-main.560} {Detect all
  abuse! toward universal abusive language detection models}.
\newblock In \emph{Proceedings of the 28th International Conference on
  Computational Linguistics}, pages 6366--6376, Barcelona, Spain (Online).
  International Committee on Computational Linguistics.

\bibitem[{Wiegand et~al.(2018)Wiegand, Siegel, and
  Ruppenhofer}]{wiegand2018overview}
Michael Wiegand, Melanie Siegel, and Josef Ruppenhofer. 2018.
\newblock Overview of the {G}erm{E}val 2018 shared task on the identification
  of offensive language.

\bibitem[{Zampieri et~al.(2019)Zampieri, Malmasi, Nakov, Rosenthal, Farra, and
  Kumar}]{zampieri-etal-2019-semeval}
Marcos Zampieri, Shervin Malmasi, Preslav Nakov, Sara Rosenthal, Noura Farra,
  and Ritesh Kumar. 2019.
\newblock \href {https://doi.org/10.18653/v1/S19-2010} {{S}em{E}val-2019 task
  6: Identifying and categorizing offensive language in social media
  ({O}ffens{E}val)}.
\newblock In \emph{Proceedings of the 13th International Workshop on Semantic
  Evaluation}, pages 75--86, Minneapolis, Minnesota, USA. Association for
  Computational Linguistics.

\bibitem[{Zhang et~al.(2018)Zhang, Robinson, and Tepper}]{zhang2018detecting}
Ziqi Zhang, David Robinson, and Jonathan Tepper. 2018.
\newblock Detecting hate speech on twitter using a convolution-gru based deep
  neural network.
\newblock In \emph{The Semantic Web: 15th International Conference, ESWC 2018,
  Heraklion, Crete, Greece, June 3--7, 2018, Proceedings 15}, pages 745--760.
  Springer.

\bibitem[{Zhou et~al.(2023)Zhou, Cabello, Cao, and
  Hershcovich}]{zhou-etal-2023-cross}
Li~Zhou, Laura Cabello, Yong Cao, and Daniel Hershcovich. 2023.
\newblock \href {https://doi.org/10.18653/v1/2023.c3nlp-1.2} {Cross-cultural
  transfer learning for {C}hinese offensive language detection}.
\newblock In \emph{Proceedings of the First Workshop on Cross-Cultural
  Considerations in NLP (C3NLP)}, pages 8--15, Dubrovnik, Croatia. Association
  for Computational Linguistics.

\bibitem[{Zhou et~al.(2020)Zhou, Yang, Liu, Liu, and Savage}]{9139953}
Yanling Zhou, Yanyan Yang, Han Liu, Xiufeng Liu, and Nick Savage. 2020.
\newblock \href {https://doi.org/10.1109/ACCESS.2020.3009244} {Deep learning
  based fusion approach for hate speech detection}.
\newblock \emph{IEEE Access}, 8:128923--128929.

\bibitem[{Zimmerman et~al.(2018)Zimmerman, Kruschwitz, and
  Fox}]{zimmerman2018improving}
Steven Zimmerman, Udo Kruschwitz, and Chris Fox. 2018.
\newblock Improving hate speech detection with deep learning ensembles.
\newblock In \emph{Proceedings of the eleventh international conference on
  language resources and evaluation (LREC 2018)}.

\bibitem[{Zoph et~al.(2016)Zoph, Yuret, May, and
  Knight}]{zoph-etal-2016-transfer}
Barret Zoph, Deniz Yuret, Jonathan May, and Kevin Knight. 2016.
\newblock \href {https://doi.org/10.18653/v1/D16-1163} {Transfer learning for
  low-resource neural machine translation}.
\newblock In \emph{Proceedings of the 2016 Conference on Empirical Methods in
  Natural Language Processing}, pages 1568--1575, Austin, Texas. Association
  for Computational Linguistics.

\end{thebibliography}

\section*{Appendix}

\begin{table*}
\centering
\scriptsize
\begin{tabular}{l l p{3cm} p{7 cm}}
\toprule
Lang & Reference & Labels & Hate definition as stated in reference \\
\midrule

BN & \cite{romim2020hatespeechdetectionbengali} &
Hate, Not hate &
A sentence that dehumanizes one or multiple persons or a community (e.g., comparing them to insects, objects, criminals, or targeting race, gender, disability). \\

UR & \cite{rizwan-etal-2020-hate} &
Hate-Offensive, Normal, Abusive, Offensive, Sexism, Religious Hate, Profane &
Language that expresses hatred, abuse, or offensive content such as profanity, vulgar or insulting language targeting individuals or groups; includes gender-based hatred and religion-based hatred. \\

HI & \cite{DBLP:journals/corr/abs-2011-03588} &
Hostile, Non-hostile, Fake News, Hate, Offensive, Defamation, Normal &
A post targeting a specific group of people based on ethnicity, religious beliefs, geographical belonging, race, etc., with malicious intentions of spreading hate or encouraging violence. \\

FA & \cite{delbari2024spanning} &
Normal, Violence, Hate, Vulgar &
Communication targeting individuals or groups with the intention of attacking, dehumanizing, inciting violence, promoting discrimination, or fostering hostility based on characteristics such as race, ethnicity, religion, gender, sexual orientation, political beliefs, social class, disability, or appearance. \\

IN & \cite{ibrohim-budi-2019-multi} &
Hate, Not hate &
Communication that disparages a person or a group based on characteristics such as race, ethnicity, gender, sexual orientation, nationality, religion, or other attributes. \\

MS & \cite{maity2023deep} &
Hate, Not hate &
Communication intended to attack the dignity of a group on the basis of characteristics such as race, gender, ethnicity, sexual orientation, nationality, religion, or other characteristics. \\

AR & \cite{mubarak-etal-2022-overview} &
Normal, Abusive, Hate &
Any communication that disparages a person or a group based on characteristics such as race, color, ethnicity, gender, sexual orientation, nationality, religion, or other characteristic. \\

TR & \cite{toraman-etal-2022-large} &
Hate, Offensive, Normal &
Tweets that target, incite violence against, threaten, or call for physical damage toward an individual or a group because of some identifying trait or characteristic. \\

KO & \cite{moon-etal-2020-beep} &
Hate, Offensive, None &
Comments that explicitly express hatred against an individual or group based on attributes such as sex, gender, sexual orientation, gender identity, age, appearance, social status, religion, disability, ethnicity, or national origin, or that severely insult or attack them. \\

ES & \cite{tonneau2024spanish} &
Hate, Not hate &
Speech that denigrates a person or multiple persons based on their membership to a group, usually defined by race, ethnicity, sexual orientation, gender identity, disability, religion, political affiliation, or views. \\

DE & \cite{tonneau-etal-2024-languages} &
Hate, Not hate, Offensive, Profanity &
Describing dehumanization or deficiencies to groups or individuals because of being members of a group (e.g., race, political opinion, sexual orientation, gender, social status, health condition). \\

FR & \cite{tonneau-etal-2024-languages} &
Abusive, Hate, Offensive, Disrespectful, Fearful, Normal &
Speech promotes the dehumanization of people who are typically from minority groups and may incite hate crimes. \\

EN & \cite{tonneau-etal-2024-languages} &
Abusive, Hate, Offensive, Disrespectful, Fearful, Normal &
Speech promotes the dehumanization of people who are typically from minority groups and may incite hate crimes. \\

\bottomrule
\end{tabular}

\caption{Comparison of hate-speech definitions and annotation schemes across languages.}
\label{tab:hate_definitions}
\end{table*}

\begin{table*}[!htbp]
\centering
\scriptsize
\setlength{\tabcolsep}{4pt}
\begin{tabular}{llcccccccccccccc}
\toprule
Method & Model & TR & UR & FA & AR & HI & BN & IN & MS & EN & ES & FR & DE & KO & ZH \\
\midrule
\multirow{4}{*}{Zero-shot}
& GPT-4o            & 71 & 45 & 55 & 64 & 67 & 53 & 65 & 56 & 82 & 75 & 77 & 75 & 60 & 67 \\
& LLaMA-3-8B-Instruct        & 70 & 52 & 45 & 72 & 70 & 58 & 64 & 51 & 82 & 79 & 76 & 74 & 49 & 66 \\
& Gemma-3-4B-Instruct  & 46 & 38 & 58 & 59 & 56 & 54 & 40 & 51 & 85 & 67 & 69 & 70 & 43 & 52 \\
& Gemma-4-4B-Instruct & 58 & 50 & 64 & 67 & 64 & 60 & 55 & 58 & 87 & 76 & 78 & 78 & 55 & 64 \\
\midrule
\multirow{4}{*}{Fine-tuning}
& LLaMA-3-8B  & \textbf{90} & \textbf{85} & 80 & 91 & \textbf{85} & \textbf{79} & 82 & 78 & 96 & 90 & 94 & 90 & 75 & 75 \\
& Gemma-3-4B  & 88 & 85 & 75 & 90 & 75 & 70 & 75 & 71 & 94 & 88 & 90 & 94 & 76 & 82 \\
& Gemma-4-4B & \textbf{92} & \textbf{88} & \textbf{82} & \textbf{93} & 84 & 78 & \textbf{84} & \textbf{79} & \textbf{97} & \textbf{92} & \textbf{95} & \textbf{96} & \textbf{81} & \textbf{86} \\

\bottomrule
\end{tabular}
\caption{F1 scores of different models for multilingual hate-speech detection.}
\label{tab:multilingual_f1}
\end{table*}

\begin{table*}
    \centering
    \includegraphics[width=0.9\textwidth]{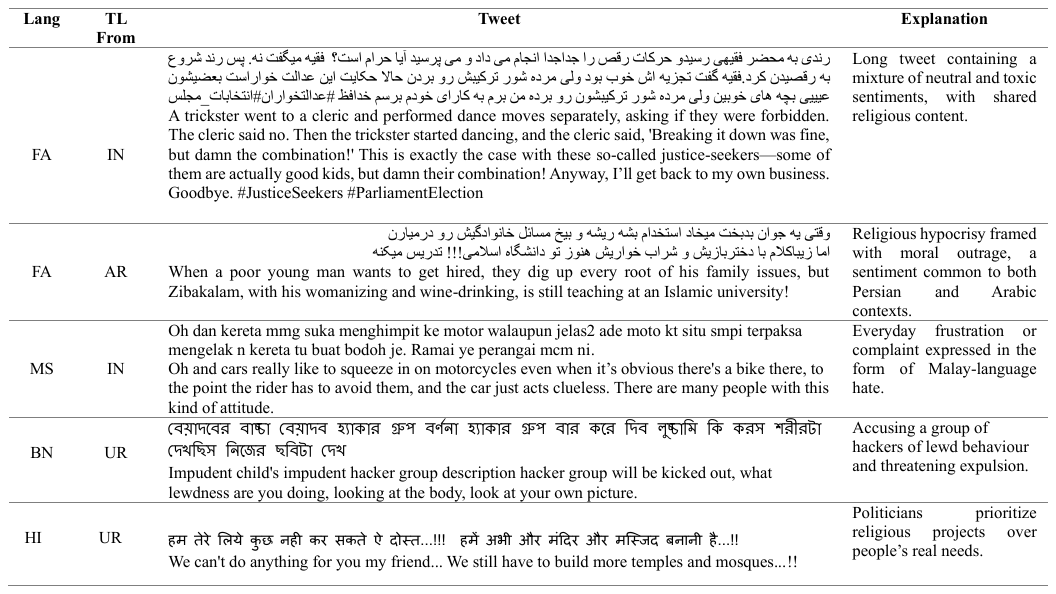} 
    \caption{Hate samples correctly predicted via transfer learning across languages.}
    \label{tab:RQ1}
\end{table*}

\begin{table*}
    \centering
    \includegraphics[width=0.8\textwidth]{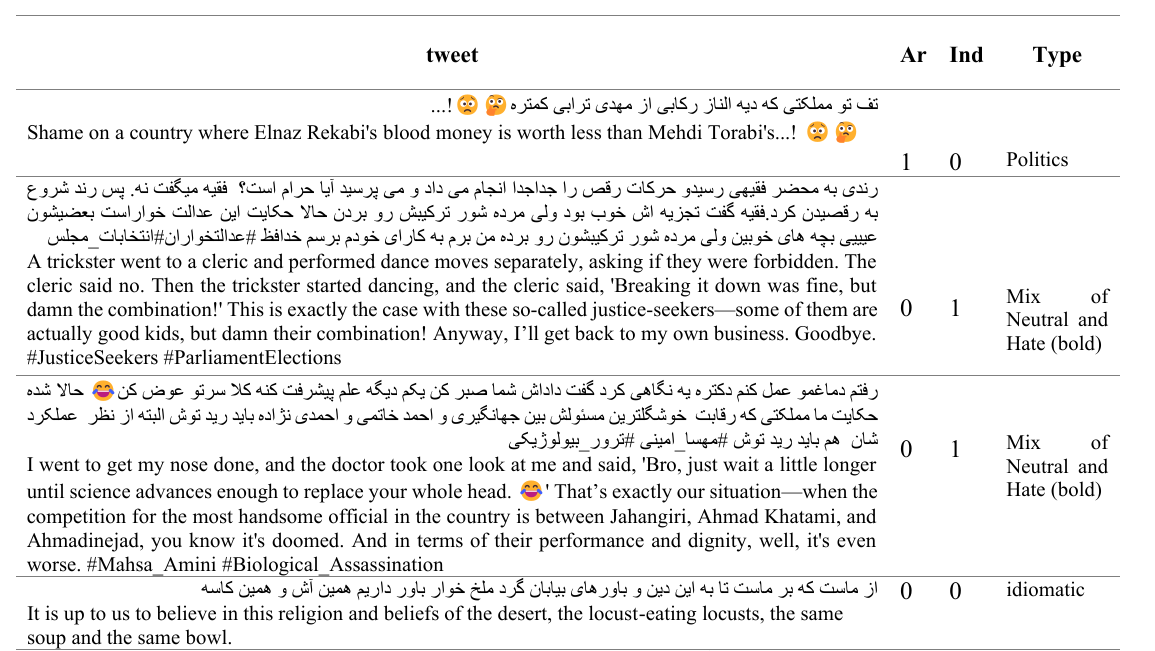}
    \caption {Case study of Persian hate-speech instances correctly predicted through transfer learning from cluster languages, or misclassified in all experimental settings. }
    \label{case-study}
\end{table*}

\begin{table*}[!htbp]
\centering
\resizebox{\textwidth}{!}{
\begin{tabular}{lcccccccccccccc}
\toprule
Strategy & TR & UR & FA & AR & HI & BN & IN & MS & EN & ES & FR & DE & KO & ZH \\
\midrule
Solo-Source TL    & 88 & 85 & 75 & 90 & 75 & 70 & 75 & 71 & 94 & 88 & 90 & 94 & 76 & 82 \\
Global-Imb        & 89 {\scriptsize[+1]} & 87 {\scriptsize[+2]} & 62 {\scriptsize[-13]} & 86 {\scriptsize[-4]} & 65 {\scriptsize[-10]} & 60 {\scriptsize[-10]} & 77 {\scriptsize[+2]} & 57 {\scriptsize[-14]} & 95 {\scriptsize[+1]} & 92 {\scriptsize[+4]} & 92 {\scriptsize[+2]} & 96 {\scriptsize[+2]} & 73 {\scriptsize[-3]} & 79 {\scriptsize[-3]} \\
Global-Bal        & 88 {\scriptsize[0]}  & 86 {\scriptsize[+1]} & 64 {\scriptsize[-11]} & 88 {\scriptsize[-2]} & 71 {\scriptsize[-4]}  & 62 {\scriptsize[-8]}  & 76 {\scriptsize[+1]} & 58 {\scriptsize[-13]} & 94 {\scriptsize[0]}  & 90 {\scriptsize[+2]} & 90 {\scriptsize[0]}  & 95 {\scriptsize[+1]} & 74 {\scriptsize[-2]} & 81 {\scriptsize[-1]} \\
Observed Cluster  & 88 {\scriptsize[0]}  & 89 {\scriptsize[+4]} & 78 {\scriptsize[+3]}  & 92 {\scriptsize[+2]} & 79 {\scriptsize[+4]}  & 73 {\scriptsize[+3]}  & 78 {\scriptsize[+3]} & 73 {\scriptsize[+2]}  & 95 {\scriptsize[+1]} & 93 {\scriptsize[+5]} & 92 {\scriptsize[+2]} & 96 {\scriptsize[+2]} & 76 {\scriptsize[0]}  & 84 {\scriptsize[+2]} \\
\bottomrule
\end{tabular}
}
\caption{\change{Performance comparison across fine-tuning strategies using Gemma-3. Bracketed numbers indicate change relative to Solo-Source TL.}{}}
\label{tab:gemma3_strategies}
\end{table*}

\begin{table*}[!htbp]
\centering
\resizebox{\textwidth}{!}{
\begin{tabular}{lcccccccccccccc}
\toprule
Strategy & TR & UR & FA & AR & HI & BN & IN & MS & EN & ES & FR & DE & KO & ZH \\
\midrule
Solo-Source TL    & 92 & 88 & 82 & 93 & 84 & 78 & 84 & 79 & 97 & 92 & 95 & 96 & 81 & 86 \\
Global-Imb        & 94 {\scriptsize[+2]} & 90 {\scriptsize[+2]} & 70 {\scriptsize[-12]} & 90 {\scriptsize[-3]} & 75 {\scriptsize[-9]}  & 69 {\scriptsize[-9]}  & 86 {\scriptsize[+2]} & 67 {\scriptsize[-12]} & 98 {\scriptsize[+1]} & 95 {\scriptsize[+3]} & 96 {\scriptsize[+1]} & 97 {\scriptsize[+1]} & 79 {\scriptsize[-2]} & 84 {\scriptsize[-2]} \\
Global-Bal        & 93 {\scriptsize[+1]} & 89 {\scriptsize[+1]} & 73 {\scriptsize[-9]}  & 92 {\scriptsize[-1]} & 81 {\scriptsize[-3]}  & 71 {\scriptsize[-7]}  & 85 {\scriptsize[+1]} & 68 {\scriptsize[-11]} & 97 {\scriptsize[0]}  & 94 {\scriptsize[+2]} & 95 {\scriptsize[0]}  & 97 {\scriptsize[+1]} & 80 {\scriptsize[-1]} & 86 {\scriptsize[0]} \\
Observed Cluster  & 92 {\scriptsize[0]}  & 92 {\scriptsize[+4]} & 85 {\scriptsize[+3]}  & 95 {\scriptsize[+2]} & 87 {\scriptsize[+3]}  & 81 {\scriptsize[+3]}  & 86 {\scriptsize[+2]} & 81 {\scriptsize[+2]}  & 98 {\scriptsize[+1]} & 96 {\scriptsize[+4]} & 96 {\scriptsize[+1]} & 98 {\scriptsize[+2]} & 81 {\scriptsize[0]}  & 88 {\scriptsize[+2]} \\
\bottomrule
\end{tabular}
}
\caption{\change{Performance comparison across fine-tuning strategies using Gemma-4. Bracketed numbers indicate change relative to Solo-Source TL.}{}}
\label{tab:gemma4_strategies}
\end{table*}

\begin{table*}[!htbp]
\centering
\small
\setlength{\tabcolsep}{4pt}
\begin{tabular}{l|cccccccccccccc}
\toprule
\textbf{Test $\downarrow$ / Train $\rightarrow$} & TR & UR & FA & AR & HI & BN & IN & MS & EN & ES & FR & DE & KO & ZH \\
\midrule
TR & \textbf{88} & 64 & 65 & 56 & 56 & 47 & 39 & 36 & 64 & 55 & 61 & 68 & 49 & 53 \\
UR & 63 & \textbf{85} & 66 & 65 & 67 & 69 & 41 & 35 & 53 & 49 & 49 & 41 & 39 & 38 \\
FA & 63 & 65 & \textbf{75} & 73 & 50 & 47 & 73 & 53 & 55 & 41 & 43 & 41 & 51 & 51 \\
AR & 54 & 67 & 67 & \textbf{90} & 51 & 47 & 49 & 45 & 55 & 47 & 43 & 49 & 53 & 47 \\
HI & 57 & 82 & 65 & 51 & \textbf{75} & 58 & 53 & 41 & 60 & 54 & 43 & 39 & 43 & 41 \\
BN & 45 & 76 & 39 & 31 & 65 & \textbf{70} & 53 & 47 & 55 & 49 & 39 & 31 & 35 & 33 \\
IN & 41 & 65 & 60 & 79 & 50 & 47 & \textbf{75} & 65 & 51 & 33 & 43 & 33 & 51 & 48 \\
MS & 35 & 51 & 53 & 60 & 39 & 39 & 75 & \textbf{71} & 47 & 41 & 51 & 47 & 47 & 51 \\
EN & 64 & 49 & 47 & 53 & 47 & 47 & 49 & 41 & \textbf{94} & 80 & 80 & 65 & 51 & 53 \\
ES & 55 & 39 & 39 & 51 & 39 & 41 & 41 & 41 & 79 & \textbf{88} & 79 & 65 & 41 & 41 \\
FR & 61 & 47 & 45 & 43 & 45 & 31 & 43 & 41 & 77 & 76 & \textbf{90} & 70 & 53 & 53 \\
DE & 67 & 38 & 41 & 47 & 41 & 39 & 43 & 37 & 69 & 67 & 69 & \textbf{94} & 41 & 45 \\
KO & 55 & 41 & 39 & 43 & 41 & 39 & 51 & 49 & 53 & 41 & 38 & 41 & \textbf{76} & 71 \\
ZH & 53 & 38 & 37 & 38 & 41 & 38 & 49 & 47 & 55 & 38 & 37 & 41 & 68 & \textbf{82} \\
\bottomrule
\end{tabular}
\caption{\change{Cross-lingual F1 scores across 14 languages using Gemma-3. Rows: test language. Columns: train language. Diagonal (matched train/test) in bold.}{}}
\label{tab:gemma3_matrix}
\end{table*}

\begin{table*}[!htbp]
\centering
\small
\setlength{\tabcolsep}{4pt}
\begin{tabular}{l|cccccccccccccc}
\toprule
\textbf{Test $\downarrow$ / Train $\rightarrow$} & TR & UR & FA & AR & HI & BN & IN & MS & EN & ES & FR & DE & KO & ZH \\
\midrule
TR & \textbf{92} & 68 & 69 & 60 & 59 & 51 & 41 & 41 & 67 & 59 & 64 & 70 & 53 & 57 \\
UR & 67 & \textbf{88} & 70 & 69 & 71 & 73 & 45 & 39 & 56 & 53 & 53 & 45 & 43 & 41 \\
FA & 69 & 70 & \textbf{82} & 75 & 55 & 51 & 75 & 56 & 60 & 47 & 47 & 47 & 55 & 55 \\
AR & 59 & 71 & 70 & \textbf{93} & 55 & 51 & 53 & 49 & 60 & 51 & 47 & 53 & 57 & 51 \\
HI & 62 & 84 & 69 & 55 & \textbf{84} & 71 & 57 & 45 & 64 & 58 & 47 & 43 & 47 & 45 \\
BN & 47 & 80 & 43 & 35 & 71 & \textbf{78} & 57 & 51 & 60 & 53 & 43 & 35 & 39 & 37 \\
IN & 45 & 69 & 71 & 81 & 53 & 51 & \textbf{84} & 69 & 55 & 37 & 47 & 37 & 55 & 51 \\
MS & 39 & 55 & 57 & 65 & 43 & 43 & 79 & \textbf{79} & 51 & 45 & 55 & 51 & 51 & 55 \\
EN & 67 & 53 & 51 & 57 & 51 & 51 & 53 & 45 & \textbf{97} & 83 & 83 & 69 & 55 & 57 \\
ES & 59 & 43 & 43 & 55 & 43 & 45 & 45 & 45 & 83 & \textbf{92} & 83 & 69 & 45 & 45 \\
FR & 65 & 51 & 49 & 47 & 49 & 35 & 47 & 45 & 81 & 80 & \textbf{95} & 73 & 57 & 57 \\
DE & 71 & 41 & 45 & 51 & 45 & 43 & 47 & 41 & 73 & 71 & 73 & \textbf{96} & 45 & 49 \\
KO & 59 & 45 & 43 & 47 & 45 & 43 & 55 & 53 & 57 & 45 & 43 & 45 & \textbf{81} & 75 \\
ZH & 57 & 41 & 41 & 43 & 45 & 41 & 53 & 51 & 59 & 41 & 41 & 45 & 71 & \textbf{86} \\
\bottomrule
\end{tabular}
\caption{\change{Cross-lingual F1 scores across 14 languages using Gemma-4. Rows: test language. Columns: train language. Diagonal (matched train/test) in bold.}{}}
\label{tab:gemma4_matrix}
\end{table*}

\begin{table*}[!htbp]
\centering
\small
\renewcommand{\arraystretch}{1.2}
\begin{tabular}{p{2.4cm}p{4.6cm}p{1.8cm}p{5.4cm}}
\toprule
\textbf{Cluster} & \textbf{Hate Train Size (per language)} & \textbf{Platform} & \textbf{Time Period} \\
\midrule
EN--FR--ES--DE & EN: 4500, FR: 3739, ES: 4731,

DE: 4503 & X & 2016--2023 \\
UR--HI--BN & UR: 4500, HI: 2222, BN: 4504 & X, FB, WA, YT & BN: 2017--2020; UR, HI: unspecified \\
FA--AR--UR--IN & FA: 1002, AR: 4503, UR: 4500, 

IN: 4502 & X & FA: 2020--2022; AR: 2016--2018; IN: 2018; UR: unspecified \\
IN--MS & IN: 4502, MS: 1701 & X & IN: 2018; MS: 2022--2023 \\
KO--ZH & KO: 1520, ZH: 2470 & X / Web & KO: 2018--2020; ZH: unspecified \\
\bottomrule
\end{tabular}
\caption{\change{Summary of dataset characteristics across the five empirically observed transfer clusters.}{}}
\label{tab:cluster_confounds}
\end{table*}

\clearpage

\section{Quantitative Analysis of Transfer Factors}
\label{app:quant_analysis}
To better understand the factors influencing cross-lingual transfer performance, we conducted a quantitative analysis examining the relationship between several dataset properties and transfer performance. For each pair of languages, the transfer performance was defined as the off-diagonal F1 score from the $14 \times 14$ transfer matrix presented in Figure \ref{fig:heatmap}. This results in 182 source–target language pairs. For each pair we computed three measurable factors: dataset size difference, temporal overlap, and linguistic similarity. The resulting dataset consists of one row per source–target pair containing the transfer F1 score and the corresponding factor values (Table~\ref{tab:transfer_prop}).

\subsection{Dataset Size Difference}

Dataset size difference measures the disparity between the number of hate instances used for training in the source and target datasets. For each language pair we computed the absolute log-ratio of the hate training sizes:
\begin{equation*}
size\_diff = |\log(N_{source}) - \log(N_{target})|
\end{equation*}

where $N_{source}$ and $N_{target}$ denote the number of hate training instances reported in Table~1. This produced 182 values corresponding to the off-diagonal entries of the transfer matrix.

\subsection{Temporal Overlap}

Temporal overlap measures how much the data collection periods of two datasets intersect. For languages where collection periods were available, we computed the overlap between the time ranges using intersection over union:

\[
temporal\_overlap =
\frac{|T_{source} \cap T_{target}|}
{|T_{source} \cup T_{target}|}
\]

where $T_{source}$ and $T_{target}$ represent the years covered by each dataset. Language pairs with missing collection dates were excluded from this calculation.

\subsection{Linguistic Similarity}

To measure linguistic proximity between languages, we used typological feature vectors derived from the URIEL/lang2vec database \cite{bjerva2019language}. Each language is represented by a vector of typological and phylogenetic features. Pairwise similarity between languages was computed using cosine similarity between these vectors:

\[
ling\_similarity = cosine(v_{source}, v_{target})
\]

where $v_{source}$ and $v_{target}$ denote the feature vectors for the source and target languages.

\subsection{Correlation Analysis}

To assess the relationship between these factors and transfer performance, we computed Spearman rank correlations \cite{spearman1961proof} between each factor and the transfer F1 scores across all language pairs. The correlation results reported in Table \ref{tab:correlation_results} indicate that linguistic similarity has the strongest association with transfer performance, followed by temporal overlap, while dataset size difference shows only a weak relationship. This trend is also visible in several high-performing language pairs with strong typological proximity. For example, transfer from Urdu to Hindi achieves the highest score in the matrix with very high linguistic similarity, while Urdu to Bengali and English to German show similarly strong performance. High transfer is also observed among closely related Romance languages, such as Spanish to French. Likewise, Indonesian transfers effectively to Malay, reflecting their near-identical linguistic structure. However, not all strong transfers align perfectly with linguistic proximity. For instance, Turkish transfers relatively well to English and German. Similarly, Arabic transfers effectively to Urdu, Indonesian to Persian despite belonging to different language families. These cases suggest that while typological similarity contributes to transfer effectiveness, it does not fully account for the patterns observed in the transfer matrix, and that broader discourse or sociocultural similarities may also facilitate cross-lingual transfer.

\section{\change{Leave-One-Language-Out Analysis}{}}
\label{app:lolo}

To verify that the gains observed in the Observed Cluster experiment are not driven by any single language within a cluster, we conduct a leave-one-language-out (LOLO) analysis on the three largest clusters. For each language $\ell$ in a cluster $C$, we remove $\ell$ from the training set, fine-tune a model on the remaining $|C|-1$ languages, and evaluate it on $\ell$. The resulting F1 score is compared against the Solo-Source TL baseline for $\ell$ and against the gain reported in the main Observed Cluster experiment (Table~\ref{fig:additional}). If the LOLO gains are comparable to the original cluster gains, this indicates that the held-out language genuinely benefits from the cluster rather than serving as a critical contributor to it.

Tables~\ref{tab:lolo_cluster1}, \ref{tab:lolo_cluster2}, and \ref{tab:lolo_cluster3} present LOLO results for the Persian-Arabic-Urdu-Indonesian, Urdu-Hindi-Bengali, and English-Spanish-French-German clusters, respectively. Across all three clusters, the LOLO gains closely match the original gains reported in Table~\ref{fig:additional}, with most differences within one F1 point. This confirms that the cluster benefit is a property of the cluster as a whole, and supports the robustness of the empirically identified groupings.

\begin{table}[!htbp]
\centering
\small
\renewcommand{\arraystretch}{1.2}
\begin{tabular}{lcccc}
\toprule
\textbf{Held-out} & \textbf{Solo F1} & \textbf{LOLO Cluster F1} & \textbf{$\Delta$ vs Solo} & \textbf{Original $\Delta$} \\
\midrule
FA & 80 $\pm$ 0.8 & 83 $\pm$ 0.5 & \textbf{+3} & +3 \\
AR & 91 $\pm$ 0.6 & 93 $\pm$ 0.4 & \textbf{+2} & +2 \\
UR & 85 $\pm$ 0.6 & 90 $\pm$ 0.4 & \textbf{+5} & +5 \\
IN & 82 $\pm$ 0.6 & 85 $\pm$ 0.4 & \textbf{+3} & +3 \\
\bottomrule
\end{tabular}
\caption{\change{LOLO results for Cluster 1: Persian-Arabic-Urdu-Indonesian. ``Original $\Delta$'' refers to the gain reported in Table~\ref{fig:additional}}.}{}
\label{tab:lolo_cluster1}
\end{table}
\FloatBarrier

\begin{table}[!htbp]
\centering
\small
\renewcommand{\arraystretch}{1.2}
\begin{tabular}{lcccc}
\toprule
\textbf{Held-out} & \textbf{Solo F1} & \textbf{LOLO Cluster F1} & \textbf{$\Delta$ vs Solo} & \textbf{Original $\Delta$} \\
\midrule
UR & 85 $\pm$ 0.6 & 88 $\pm$ 0.5 & \textbf{+3} & +5 \\
HI & 80 $\pm$ 0.7 & 83 $\pm$ 0.5 & \textbf{+3} & +5 \\
BN & 70 $\pm$ 0.7 & 73 $\pm$ 0.5 & \textbf{+3} & +4 \\
\bottomrule
\end{tabular}
\caption{\change{LOLO results for Cluster 2: Urdu--Hindi--Bengali.}{}}
\label{tab:lolo_cluster2}
\end{table}
\FloatBarrier

\begin{table}[!htbp]
\centering
\small
\renewcommand{\arraystretch}{1.2}
\begin{tabular}{lcccc}
\toprule
\textbf{Held-out} & \textbf{Solo F1} & \textbf{LOLO Cluster F1} & \textbf{$\Delta$ vs Solo} & \textbf{Original $\Delta$} \\
\midrule
EN & 96 $\pm$ 0.5 & 96 $\pm$ 0.4 & \textbf{0}  & +1 \\
ES & 90 $\pm$ 0.6 & 94 $\pm$ 0.4 & \textbf{+4} & +6 \\
FR & 94 $\pm$ 0.5 & 95 $\pm$ 0.4 & \textbf{+1} & +2 \\
DE & 90 $\pm$ 0.6 & 93 $\pm$ 0.4 & \textbf{+3} & +5 \\
\bottomrule
\end{tabular}
\caption{\change{LOLO results for Cluster 3: English--Spanish--French--German.}{}}
\label{tab:lolo_cluster3}
\end{table}
\FloatBarrier

\begin{table*}
\centering
\scriptsize
\setlength{\tabcolsep}{6pt}
\renewcommand{\arraystretch}{0.9}
\begin{tabular}{llllll}
\toprule
source & target & F1 & size\_diff & ling\_similarity & temporal\_overlap \\
\midrule
TR & FA & \textbf{66} & 1.5 & 0.7 & 0.29 \\
TR & MS & 39 & 0.97 & 0.46 & 0.14 \\
TR & ES & 57 & 0.05 & 0.47 & 1 \\
TR & FR & 62 & 0.19 & 0.48 & 1 \\
\midrule
FA & TR & 48 & 1.5 & 0.7 & 0.29 \\
FA & AR & 63 & 1.5 & 0.78 & 0 \\
FA & BN & 43 & 1.5 & 0.7 & 0 \\
FA & IN & \textbf{67} & 1.5 & 0.58 & 0 \\
FA & KO & 40 & 0.42 & 0.4 & 0 \\
\midrule
AR & FA & \textbf{80} & 1.5 & 0.78 & 0 \\
AR & MS & 53 & 0.97 & 0.56 & 0 \\
AR & ES & 55 & 0.05 & 0.58 & 0.29 \\
AR & KO & 44 & 1.09 & 0.38 & 0 \\
\midrule
BN & MS & 43 & 0.97 & 0.61 & 0 \\
BN & ES & \textbf{61} & 0.05 & 0.61 & 0.43 \\
\midrule
IN & MS & \textbf{78} & 0.97 & 0.92 & 0 \\
IN & ES & 57 & 0.05 & 0.63 & 0.29 \\
IN & FR & 57 & 0.19 & 0.64 & 0.29 \\
IN & KO & 62 & 1.09 & 0.42 & 0 \\
\midrule
MS & IN & \textbf{63} & 0.97 & 0.92 & 0 \\
MS & EN & 50 & 0.97 & 0.66 & 0.14 \\
MS & ES & 47 & 1.02 & 0.63 & 0.14 \\
MS & FR & 55 & 0.79 & 0.64 & 0.14 \\
\midrule
EN & FA & 56 & 1.5 & 0.63 & 0.29 \\
EN & FR & \textbf{78} & 0.19 & 0.88 & 1 \\
EN & KO & 73 & 1.09 & 0.45 & 0.29 \\
\midrule
ES & TR & \textbf{65} & 0.05 & 0.47 & 1 \\
ES & FA & 50 & 1.55 & 0.61 & 0.29 \\
ES & BN & 54 & 0.05 & 0.61 & 0.43 \\
ES & IN & 47 & 0.05 & 0.63 & 0.29 \\
\midrule
FR & TR & 54 & 0.19 & 0.48 & 1 \\
FR & BN & \textbf{58} & 0.19 & 0.62 & 0.43 \\
FR & IN & 40 & 0.19 & 0.64 & 0.29 \\
FR & MS & 49 & 0.79 & 0.64 & 0.14 \\
\midrule
DE & FA & 45 & 1.5 & 0.64 & 0.29 \\
DE & ES & \textbf{68} & 0.05 & 0.88 & 1 \\
\midrule
KO & TR & \textbf{58} & 0.11 & 0.43 & 0 \\
KO & IN & 44 & 1.09 & 0.42 & 0 \\
KO & MS & 58 & 0.11 & 0.43 & 0 \\
KO & ES & 43 & 1.14 & 0.44 & 0.29 \\
\bottomrule
\end{tabular}
\caption{Cross-lingual transfer cases with corresponding dataset properties for a representative sample.}
\label{tab:transfer_prop}
\end{table*}

\end{document}